\pdfoutput=1
\documentclass{article}

\usepackage{iclr2027_conference,times}

\newcommand{\KL}{\mathrm{KL}}

\newcommand{\softmax}{\operatorname{softmax}}
\newcommand{\Lgrpo}{\mathcal L_{\mathrm{GRPO}}}
\newcommand{\Lref}{\mathcal L_{\mathrm{ref}}}
\newcommand{\Levi}{\mathcal L_{\mathrm{evi}}}
\newcommand{\Lverpo}{\mathcal L_{\mathrm{VERPO}}}
\newcommand{\V}{\mathcal V}

\usepackage[hypertexnames=false]{hyperref}
\usepackage{url}
\usepackage{booktabs}
\usepackage{pifont}
\usepackage{amsmath,amssymb,mathtools,bm}
\usepackage{microtype}
\usepackage{graphicx}
\usepackage{xcolor}
\definecolor{githubpink}{RGB}{255,0,128}
\usepackage{tabularx}
\usepackage{array}
\usepackage{enumitem}
\usepackage{algorithm}
\usepackage{algorithmic}
\usepackage[most]{tcolorbox}
\usepackage{needspace}
\usepackage{newtxtext}
\usepackage{newtxmath}
\usepackage{wrapfig}
\usepackage{tikz}
\usepackage{caption}
\usepackage{subcaption}
\newcommand{\papertitle}{
VERPO: Verified Evidence Regularized Policy Optimization}

\title{\papertitle}

\author{%
\begin{minipage}{\dimexpr\textwidth-2\tabcolsep\relax}
\raggedright\normalfont
\hbox to \linewidth{%
\mbox{\textbf{Haijiang Li}$^{1\ast}$}\hfill%
\mbox{\textbf{Chengyu Lv}$^{2\ast}$}\hfill%
\mbox{\textbf{Yi Zhang}$^{3\ast}$}\hfill%
\mbox{\textbf{Rui Qian}$^{4\ast}$}\hfill%
\mbox{\textbf{Zhibing Zhang}$^{5\ast}$}\hfill%
\mbox{\textbf{Xiangqing Shen}$^{5}$}}%
\vspace{2pt}
\hbox to \linewidth{%
\mbox{\textbf{Junjie Yang}$^{3}$}\hfill%
\mbox{\textbf{Yuchen Zhang}$^{1}$}\hfill%
\mbox{\textbf{Wenyuan Jiang}$^{6}$}\hfill%
\mbox{\textbf{Hanqing Hu}$^{7}$}\hfill%
\mbox{\textbf{Cangqi Zhou}$^{1\ddagger}$}}%
\vspace{5pt}
{\small
$^{1}$Nanjing University of Science and Technology\quad
$^{2}$Tongji University\quad $^{3}$Independent\quad $^{4}$Fudan University\\
$^{5}$Nanjing University\quad $^{6}$D-INFK, ETH Z\"urich\quad
$^{7}$Shanghai Jiaotong University}
\endgraf\vspace{4pt}
{\small
\texttt{\{\href{mailto:haijiangli23@gmail.com}{haijiangli23},\,\href{mailto:qiianruii@gmail.com}{qiianruii}\}@gmail.com},\quad
\href{mailto:cqzhou@njust.edu.cn}{\texttt{cqzhou@njust.edu.cn}}\endgraf}
\end{minipage}%
}

\newcommand{\correspondencefootnote}{%
\begingroup
\renewcommand{\thefootnote}{\fnsymbol{footnote}}%
\footnotetext[1]{Equal contribution.\qquad $^{\ddagger}$Corresponding author: Cangqi Zhou.}%
\endgroup
}
\hypersetup{pdfauthor={Haijiang Li, Chengyu Lv, Yi Zhang, Rui Qian, Zhibing Zhang, Xiangqing Shen, Junjie Yang, Yuchen Zhang, Wenyuan Jiang, Hanqing Hu, Cangqi Zhou}}

\newcommand{\method}{\textsc{VERPO}}
\newcommand{\FKL}{\mathrm{FKL}}
\newcommand{\RKL}{\mathrm{RKL}}
\newcommand{\FEC}{\mathrm{FEC}}

\newcommand{\sg}{\operatorname{sg}}
\newcommand{\clip}{\operatorname{clip}}
\newcommand{\diag}{\operatorname{diag}}

\newcolumntype{Y}{>{\raggedright\arraybackslash}X}
\definecolor{promptback}{HTML}{F7F7F7}
\definecolor{promptframe}{HTML}{555555}
\newtcblisting[auto counter,number within=section]{promptbox}[2][]{%
  enhanced,
  breakable,
  listing only,
  listing engine=listings,
  colback=promptback,
  colframe=promptframe,
  boxrule=0.5pt,
  arc=1pt,
  left=1.5mm,
  right=1.5mm,
  top=1mm,
  bottom=1mm,
  title={Listing~\thetcbcounter: #2},
  fonttitle=\bfseries\small,
  colbacktitle=promptframe,
  attach boxed title to top left={xshift=1.5mm,yshift=-2mm},
  boxed title style={arc=1pt,boxrule=0pt},
  label={#1},
  listing options={
    basicstyle=\ttfamily\footnotesize,
    breaklines=true,
    breakatwhitespace=false,
    columns=fullflexible,
    keepspaces=true,
    showstringspaces=false
  }
}

\iclrfinalcopy
\usepackage{etoolbox}
\makeatletter
\patchcmd{\@maketitle}
  {Published as a conference paper at ICLR 2027}
  {Preprint}
  {}{\PackageError{arxiv-preprint}{Unable to set the preprint header}{}}
\patchcmd{\@maketitle}
  {\LARGE\sc}
  {\fontsize{15}{18}\selectfont\scshape}
  {}{\PackageError{arxiv-preprint}{Unable to set the title size}{}}
\patchcmd{\@maketitle}
  {\@title\par}
  {\@title\par\vspace{5pt}}
  {}{\PackageError{arxiv-preprint}{Unable to set the title-author spacing}{}}
\makeatother
\hypersetup{hidelinks,pdftitle={VERPO: Verified Evidence Regularized Policy Optimization}}

\usepackage{amsmath}
\usepackage{algorithm}
\usepackage{algorithmic}
\usepackage{xcolor}

\definecolor{verpoRed}{HTML}{F2385A}
\definecolor{verpoOrange}{HTML}{FF781F}
\definecolor{verpoBlue}{HTML}{168CFF}
\definecolor{verpoGray}{HTML}{555555}

\newcommand{\VERPOhl}[2]{%
  \begingroup%
  \setlength{\fboxsep}{1pt}%
  \colorbox{#1!18}{$\mathstrut #2$}%
  \endgroup%
}

\newcommand{\VERPOcomment}[2][verpoGray]{%
  \unskip\hspace{0.6em}\hfill
  \mbox{\footnotesize
    \raisebox{-0.15ex}[0pt][0pt]{%
      \textcolor{#1}{\scalebox{1.7}{$\triangleright$}}%
    }\,%
    \textcolor{verpoGray}{#2}}%
}
\begin{document}
\begingroup
\setlength{\parskip}{4pt}
\raggedbottom

% \vspace{-2.4\baselineskip}

\maketitle
\correspondencefootnote

% Leave a little more space between the email line and Abstract.
\vspace{\dimexpr-1.6\baselineskip+1.5pt\relax}

\begingroup
\setlength{\topsep}{2pt}
\begin{abstract}
Verifiable rewards improve language models through reliable task-level feedback, but methods based on Group Relative Policy Optimization (GRPO) apply a sequence-level advantage uniformly across all tokens. This coarse credit assignment reinforces or penalizes entire responses without identifying which local decisions to preserve, reinforce, or revise. Conversely, evidence-conditioned self-distillation provides denser token-level supervision, yet teacher imitation can transfer stylistic artifacts and miscalibrated confidence that destabilize training when misaligned with task success. 
We introduce VERPO, which converts evidence-conditioned guidance into reward-aligned token-level credit assignment while retaining the outcome objective. VERPO decomposes teacher guidance into an evidence-free reference term and signed, evidence-induced corrections at each token. A stopped controller combines selective acceptance, token-wise localization, and cost-aware scaling by balancing alignment with the local GRPO update direction against Fisher movement cost. Furthermore, we introduce Fisher Evidence Contrast (FEC), which attenuates nuisance shifts along an estimated evidence-presence direction through a regularized projection. 
Across five scientific reasoning and tool-use tasks, VERPO prevents optimization collapse and consistently achieves the highest multi-task average across model backbones, yielding marked improvements particularly on smaller models over strong baselines. Qualitative diagnostics confirm that token acceptance selectively targets reasoning bottlenecks consistent with local reward alignment and Fisher movement cost. Code is available at \href{https://github.com/hamsterjiang23/VERPO}{\textcolor{githubpink}{GitHub}}.
\end{abstract}

\endgroup
\vspace{\dimexpr-1.8\baselineskip+1.5pt\relax}
\section{Introduction}

Outcome-based policy optimization with verifiable rewards has become a central approach for improving the reasoning behavior of language models~\citep{deepseekai2025deepseekr1,yu2025dapo}. In reinforcement learning with verifiable rewards (RLVR), task success can be checked automatically, allowing policy updates to be grounded in reliable outcome-level feedback. Group Relative Policy Optimization (GRPO)~\citep{shao2024deepseekmath} derives group-relative advantages from multiple on-policy rollouts and uses them to reinforce or suppress generated responses. In Fig.~\ref{fig:motivation}(a), GRPO~\citeyearpar{shao2024deepseekmath} resembles trial-and-error exploration: the reward is reliable, but it is returned only after a full response is completed. Consequently, the same sequence-level advantage is applied uniformly across response tokens, leaving the model without explicit information about which local decisions should be preserved, reinforced, or revised~\citep{hubotter2026sdpo}.

Evidence-conditioned self-distillation offers a natural way to make this feedback denser~\citep{zhao2026opsd,hubotter2026sdpo}. A self-teacher can replay sampled trajectories with privileged evidence, such as training-only feedback, reference information, or external signals, and then provide token-level guidance to a student that will not receive such evidence at deployment time. In Fig.~\ref{fig:motivation}(b), evidence can point to local correction opportunities along a rollout trajectory rather than waiting until the final answer is judged. However, we observe that relatively little work has examined whether an evidence-induced teacher change is actually a reward-aligned correction. A teacher conditioned on privileged evidence can induce shifts  not only task-relevant decisions, but also formatting, verbosity, confidence, or reasoning style. Naively imitating such a teacher can therefore transfer evidence-induced artifacts and destabilize learning when these changes are not aligned with verifiable task success~\citep{armandpour2026unmasking,pibias2026break}.

Table~\ref{tab:paradigm-comparison} summarizes this gap. GRPO~\citeyearpar{shao2024deepseekmath} preserves a direct reward objective, but lacks dense token-level guidance and cannot decide which local decisions should receive credit or correction.
% Fill page 1 before starting the unchanged wrapfigure on page 2.
\par\newpage
\begin{wrapfigure}{r}{0.45\textwidth}
    \centering

    % ==================================================
    % 上方：图片及其 caption
    % ==================================================
    \begin{minipage}{\linewidth}
        \centering

        \captionsetup{
            skip=3pt,
            position=bottom
        }

        % The PDF already contains the panel frames and vector headings.
        \includegraphics[width=\linewidth]{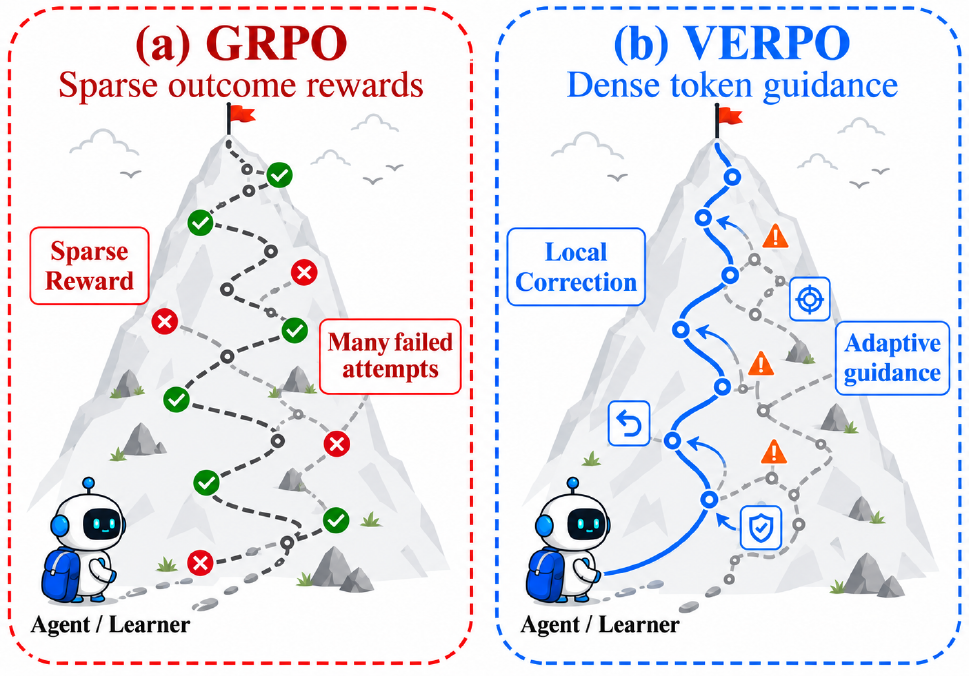}

        \caption{GRPO~\citeyearpar{shao2024deepseekmath} vs. VERPO (Ours).
        GRPO~\citeyearpar{shao2024deepseekmath} uses sparse sequence-level rewards
        for trial-and-error exploration in (a), whereas the proposed VERPO
        converts evidence-conditioned guidance into reward-aligned token-level
        credit assignment in (b).}
        \label{fig:motivation}
    \end{minipage}

    % 图片 caption 与表格 caption 之间的距离
    \par\vspace{0.6em}

    % ==================================================
    % 下方：表格及其 caption
    % ==================================================
    \begin{minipage}{\linewidth}
        \centering
        \small

        % 局部使用 table 类型，保留独立表格编号
        \captionsetup{
            type*=table,
            font=small,
            skip=3pt,
            position=top
        }

        \caption{Token-level credit mechanisms.
        DRO: direct reward objective.
        DTG: dense token guidance/credit modulation.
        SA: Selective Acceptance via sample/token
        gating beyond evidence availability.
        TL: Token-wise Localization via explicit gating.
        CS: Cost-aware Scaling by local policy movement.}
        \label{tab:paradigm-comparison}

        % 表格行距与单元格左右留白
        \renewcommand{\arraystretch}{1.05}
        \setlength{\tabcolsep}{3pt}

        % 对勾与叉号
        \newcommand{\capyes}{{\large\textcolor{black}{\ding{51}}}}
        \newcommand{\capno}{{\large\textcolor{red}{\ding{55}}}}

        % The table fills the wrapfigure width.
        % Method 使用自然宽度，后五列均分剩余宽度
        \begin{tabularx}{\linewidth}{
            @{}l|*{5}{>{\centering\arraybackslash}X}@{}
        }
            \toprule
            Method
                & DRO
                & DTG
                & SA
                & TL
                & CS \\
            \midrule

            GRPO~\citeyearpar{shao2024deepseekmath}
                & \capyes
                & \capno
                & \capno
                & \capno
                & \capno \\

            SDPO~\citeyearpar{hubotter2026sdpo}
                & \capno
                & \capyes
                & \capno
                & \capno
                & \capno \\

            SRPO~\citeyearpar{li2026srpo}
                & \capyes
                & \capyes
                & \capyes
                & \capno
                & \capno \\

            RLSD~\citeyearpar{yang2026rlsd}
                & \capyes
                & \capyes
                & \capno
                & \capno
                & \capno \\

            RLCSD~\citeyearpar{pan2026rlcsd}
                & \capyes
                & \capyes
                & \capyes
                & \capyes
                & \capno \\

            VERPO (Ours)
                & \capyes
                & \capyes
                & \capyes
                & \capyes
                & \capyes \\

            \bottomrule
        \end{tabularx}
    \end{minipage}

    % 沿用原来的底部压缩量
    % 若编译后正文与表格重叠，改为 -1\baselineskip 或删除此行
    \par\vspace{-2\baselineskip}
\end{wrapfigure}

% 后面直接接需要环绕的正文段落

\noindent
Self-distillation methods such as SDPO~\citep{hubotter2026sdpo} provide dense token guidance, but do not by themselves anchor each local teacher preference to verifiable reward improvement. What is missing is not merely another weighted combination of an RL loss and a distillation loss. Rather, the central problem is how to convert evidence-conditioned guidance into reward-aligned token-level credit for outcome-based policy optimization, while accounting for the cost of changing the policy. This requires deciding which evidence-induced corrections should be accepted, where along the response they should act, and how strongly they should be scaled. This leads us to ask: \emph{How can evidence-conditioned guidance be converted into reward-aligned token-level credit while accounting for the cost of changing the policy?}

Bearing this in mind, we introduce \textbf{VERPO}, a \textbf{V}erified \textbf{E}vidence-\textbf{R}egularized \textbf{P}olicy \textbf{O}ptimization framework. VERPO treats evidence as a proposal for policy correction rather than as a replacement for the outcome objective~\citep{shao2024deepseekmath}. It decomposes teacher guidance into two components: an evidence-free reference term that preserves the behavior available without privileged evidence, and a signed evidence-induced correction that proposes how the policy should move at each token. This separation allows the method to distinguish reference preservation from evidence-driven change, instead of collapsing both into a single imitation target.

To this end, VERPO uses a stopped token-wise ZPD controller, named after the zone of proximal development~\citep{vygotsky1978mind}. This concept distinguishes independent capabilities from those attainable with guidance. Analogously, the controller scales corrections by their alignment with the local GRPO~\citeyearpar{shao2024deepseekmath} update direction and Fisher movement cost. Guidance is strongest for reward-aligned corrections with modest policy movement, and weakest for costly or misaligned corrections. In Table~\ref{tab:paradigm-comparison}, VERPO frames reward-aligned token-level credit assignment through three missing mechanisms: 1) selective acceptance, 2) token-wise localization, and 3) cost-aware scaling of evidence-induced corrections. To construct evidence directions, VERPO supports multiple teacher comparisons. The Fix construction compares an evidence-conditioned teacher with an evidence-free teacher, while the contrastive construction compares teacher distributions under correct and incorrect evidence. To reduce nuisance shifts caused by the mere presence of privileged evidence, we introduce Fisher Evidence Contrast (FEC), which projects out the corresponding evidence-presence component. These correction directions can be used through two complementary update paths: they can weight the evidence-distillation loss, modulate the token-level policy advantage, or enter both paths through the same stopped controller. In summary, our contributions are threefold:
\vspace{0.2cm}
\begin{itemize}[
    leftmargin=0.5em,
    labelsep=0.2em,
    topsep=-\parskip,
    itemsep=2pt,
    parsep=0pt,
    partopsep=0pt
]
    \item We formulate evidence-conditioned self-distillation for RLVR
    as a reward-aligned token-level credit assignment problem.
    This formulation separates evidence-free reference preservation
    from signed evidence-induced policy correction, clarifying why
    dense teacher guidance should be treated as a local correction
    proposal rather than a direct imitation target.

    \item We present VERPO, which uses a stopped token-wise ZPD controller
    to accept and scale each correction according to local reward
    alignment and Fisher movement cost. It further supports multiple
    evidence-direction constructions, including Fix, contrastive
    evidence correction, and Fisher Evidence Contrast, which removes
    nuisance shifts caused by evidence presence.

    \item We conduct extensive experiments on a five-task suite
    spanning scientific reasoning and tool use~\citep{feng2024sciknoweval,tang2023toolalpaca} across three model
    backbones. Diagnostic analyses further reveal selective,
    task-dependent correction behavior, showing that evidence is
    most useful when it is locally reward-aligned and incurs a small
    Fisher movement cost.
\end{itemize}

\section{Related Work}
\label{sec:related-work}

\paragraph{Outcome Rewards and Token-level Supervision.}
GRPO~\citep{shao2024deepseekmath} learns from verifiable outcomes using
sequence-level advantages shared across tokens. Process supervision, including
Let's Verify Step by Step~\citep{lightman2023lets} and
Math-Shepherd~\citep{wang2024mathshepherd}, supplies finer-grained targets
through step verification~\citep{setlur2025rewarding}. Math-Shepherd~\citep{wang2024mathshepherd} constructs step labels automatically,
reducing dependence on human annotations while retaining an explicit process
supervision stage. This distinction separates the source of supervision from
the granularity at which it is applied: a reliable final outcome does not itself
identify which intermediate decisions deserve correction~\citep{hubotter2026sdpo}.
VERPO retains outcome-based optimization and uses
Teacher replay to propose local corrections without a learned process verifier.

\paragraph{Self-distillation and Reward Alignment.}
On-policy distillation~\citep{agarwal2023onpolicy,gu2023minillm} trains on student-generated
trajectories with teacher feedback, addressing the mismatch between fixed
training sequences and the student's own generations. Privileged
self-distillation instead obtains an informative Teacher view by enriching
the context available during replay~\citep{lu2026sdar}.
OPSD~\citep{zhao2026opsd} and SDPO~\citep{hubotter2026sdpo} obtain dense
supervision from self-teachers conditioned on privileged solutions or feedback.
SRPO~\citep{li2026srpo} routes samples between reward optimization and
self-distillation, while RLSD~\citep{yang2026rlsd} modulates distillation
with reward-directed signals. These mechanisms motivate selective guidance,
but do not determine token-wise acceptance by jointly measuring local reward
alignment and policy movement cost. VERPO uses a stopped ZPD weight to make
this decision for each eligible token. It keeps the evidence-free reference
term active when an evidence correction is rejected, separating preservation
of the base behavior from acceptance of a proposed change.

\paragraph{Contrastive and Selective Privileged Supervision.}
Contrastive methods separate task evidence from contextual shifts.
RLCSD~\citep{pan2026rlcsd} contrasts correct and incorrect hints to suppress
style drift. CEPO~\citep{heakl2026cepo} uses correct-versus-wrong evidence
for token credit, while OCSD~\citep{yang2026ocsd} contrasts matched replay
views to discount shared scaffold effects.
Selective methods instead restrict where supervision applies:
EDGE-OPD~\citep{lazaridis2026edgeopd} masks tokens by positive evidence,
and ROSD~\citep{zhao2026rosd} localizes correction to erroneous spans.
Distillation studies identify reference-trajectory bias in privileged
Teachers~\citep{pibias2026break} and weaker gradient alignment on correct
than incorrect rollouts~\citep{armandpour2026unmasking}.
SA-OPD~\citep{jiang2026saopd} filters signals that combine weak input
grounding with extreme divergence.
These approaches address evidence relevance, correction location, and
signal reliability. VERPO additionally separates evidence-free reference
restoration from signed correction, uses FEC to attenuate a measured
evidence-presence component under the Fisher metric, and accepts token-wise
corrections according to local reward alignment and movement cost.
Appendix~\ref{sec:app-related-work} provides the extended comparison.

\section{
VERPO:
\underline{V}erified 
\underline{E}vidence
\underline{R}egularized
\underline{P}olicy
\underline{O}ptimization
}
\label{sec:method}

\subsection{Overview of VERPO}
\label{sec:overview}

VERPO separates signed Teacher corrections from token-wise acceptance.
ZPD scales each correction by local reward alignment and Fisher movement cost,
while an evidence-free reference anchors the Actor's behavior.
We use forward-KL geometry, denoted FKL. Appendix~\ref{sec:app-rkl} derives
the reverse-KL counterpart, denoted RKL.
Figure~\ref{fig:method} summarizes four training stages:

\begin{itemize}[leftmargin=0.5em,labelsep=0.2em,partopsep=0pt]
    \item \textbf{Stage 1: Rollout Sampling and Outcome Signal.}
    The Actor samples response groups, and verifier rewards yield
    the GRPO~\citep{shao2024deepseekmath} advantage $\widehat A_i$.

    \item \textbf{Stage 2: Token-Level Evidence Correction.}
    Teacher replay under different evidence contexts yields
    Fix~\citep{hubotter2026sdpo}, CTR \citep{pan2026rlcsd}, or FEC
    corrections, defined in Section~\ref{sec:fec}.

    \item \textbf{Stage 3: ZPD Weight Estimation.}
    Local alignment and Fisher cost determine each eligible token's
    stopped weight $w_{i,t}^{*}$, derived in Section~\ref{sec:zpd}.

    \item \textbf{Stage 4: Two-Path Policy Update.}
    The weight either scales the evidence correction or modulates
    the GRPO~\citep{shao2024deepseekmath} advantage.
    Section~\ref{sec:two-paths} defines both update paths.
\end{itemize}

\begin{figure*}[t]
    \centering
    \includegraphics[width=\textwidth]{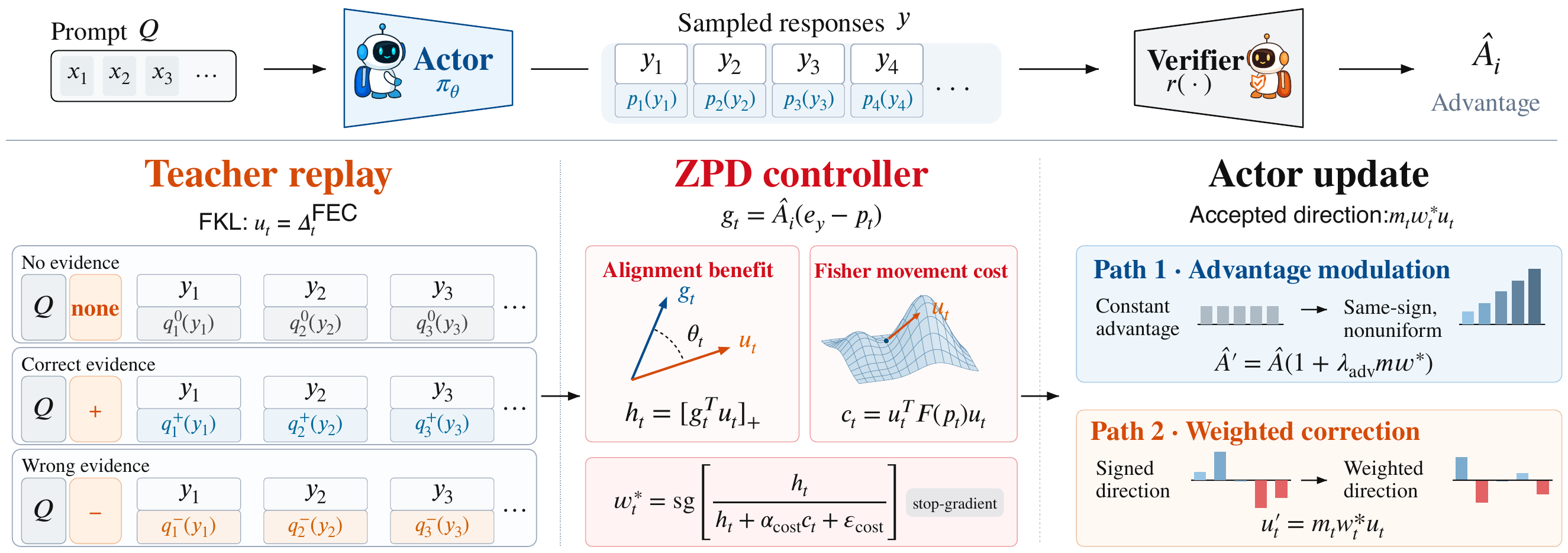}
\vspace{0.1pt}
\caption{Overview of VERPO.
Teacher replay constructs FEC correction directions.
The ZPD controller balances alignment benefit against Fisher movement cost
to determine token-wise acceptance.
Path~1 modulates advantages without changing their signs;
Path~2 applies weighted corrections.}

    \label{fig:method}
\end{figure*}

\subsection{FKL Evidence Direction and OPSD Decomposition}
\label{sec:directions}

We separate reference restoration from evidence correction using base and
enriched Teacher endpoints $q_b,q_e$, defined in Section~\ref{sec:fec}.
The Actor $p_{i,t}=\pi_\theta(\cdot\mid x_i,y_{i,<t})$ uses the original
prompt and sampled prefix. For one Teacher comparison, define
$\Delta=q_e-q_b$ and $q_w^{\FKL}=q_b+w\Delta$ for $0\leq w\leq1$.
Here $\Delta$ specifies the signed evidence correction, while $w$ controls
how far the Teacher target moves from $q_b$ toward $q_e$. The path preserves
normalization and nonnegativity. Holding the Teacher endpoints and $w$ fixed,
and assuming positive Actor probability on the Teacher support, we expand
the forward KL as
\begin{equation}
\KL(q_w^{\FKL}\Vert p)
=\KL(q_b\Vert p)
-w\sum_{v\in\V}\Delta(v)\log p(v)
+C(q_w^{\FKL},q_b),
\label{eq:main-fkl-decomposition}
\end{equation}
The first term restores $q_b$, the second contributes $w\Delta$ to the
negative logit gradient, and $C$ has no Actor gradient.
Appendix~\ref{sec:app-fkl} derives both in
Equations~\eqref{eq:app-fkl-decomposition} and~\eqref{eq:app-fkl-gradient}.
We therefore restore an independent evidence-free reference
$q_{i,t}^{\mathrm{ref}}$ through
\begin{equation}
\mathcal L_{\mathrm{ref}}^{\FKL}
=\frac1N\sum_{i,t}a_{i,t}
\KL(q_{i,t}^{\mathrm{ref}}\Vert p_{i,t}),
\label{eq:main-fkl-reference-loss}
\end{equation}
Here $a_{i,t}$ selects valid response tokens and $N$ counts them.
This loss remains active independently of evidence mask $m_{i,t}$ and
stopped acceptance $w_{i,t}$, which control the correction. Thus, rejecting
an evidence-induced change does not disable reference restoration.
The evidence loss is
\begin{equation}
\mathcal L_{\mathrm{OPSD}}^{\FKL}(w)
=-\frac1{Z_{\mathrm{evi}}}
\sum_{i,t}m_{i,t}w_{i,t}
\sum_{v\in\V}\Delta_{i,t}(v)\log p_{i,t}(v).
\label{eq:main-fkl-opsd-loss}
\end{equation}
$Z_{\mathrm{evi}}$ counts eligible tokens with a floor of one, and zero
weights remove corrections. These losses use the full vocabulary.
Appendix~\ref{sec:app-method-hyperparameters} details finite-support computation.

\subsection{Token-Wise ZPD Controller}
\label{sec:zpd}
To derive acceptance, let $\theta_{\mathrm{base}}$ denote a hypothetical
Actor state after GRPO~\citep{shao2024deepseekmath} and reference updates.
Training instead uses a joint loss, with $\bar\theta$ denoting the EMA Teacher.
For logits $z_{i,t}$ and Jacobian $G_{i,t}=\partial z_{i,t}/\partial\theta$,
the FKL direction $u_{i,t}=\Delta_{i,t}$ induces
$d_{i,t}=G_{i,t}^{\top}u_{i,t}$. At step size $\eta>0$, ideal acceptance is
\begin{equation}
w_{i,t}^{\mathrm{ideal}}
=\arg\max_{0\leq w\leq1}
J_{\mathrm{RL}}\!\left(\theta_{\mathrm{base}}+\eta w d_{i,t}\right).
\label{eq:main-zpd-ideal}
\end{equation}
For twice-differentiable $J_{\mathrm{RL}}$, a second-order expansion gives
\begin{equation}
J_{\mathrm{RL}}(\theta_{\mathrm{base}}+\eta w d_{i,t})
-J_{\mathrm{RL}}(\theta_{\mathrm{base}})\approx\eta w\,\nabla J_{\mathrm{RL}}(\theta_{\mathrm{base}})^\top d_{i,t}
+\tfrac12\eta^2w^2\,d_{i,t}^{\top}
\nabla^2J_{\mathrm{RL}}(\theta_{\mathrm{base}})d_{i,t}.
\label{eq:main-zpd-taylor}
\end{equation}
The linear term measures reward alignment, but the quadratic term penalizes
movement only under negative directional curvature. We use local alignment
and nonnegative Fisher cost proxies instead. With sampled-token one-hot
vector $e_{y_{i,t}}$, $[x]_+=\max(x,0)$, and
$F(p)=\diag(p)-pp^\top$, define
\begin{equation}
\boxed{
h_{i,t}
=\left[
\widehat A_i(e_{y_{i,t}}-p_{i,t})^\top u_{i,t}
\right]_+,
\qquad
c_{i,t}=u_{i,t}^{\top}F(p_{i,t})u_{i,t}.}
\label{eq:zpd-proxies-main}
\end{equation}

\noindent
\begin{minipage}[c]{0.75\columnwidth}
\paragraph{Selective Acceptance.}
$h_{i,t}$ measures positive alignment with the unclipped local
GRPO~\citep{shao2024deepseekmath} score direction, not verified improvement.
For nonzero $\widehat A_i(e_{y_{i,t}}-p_{i,t})$ and $u_{i,t}$,
an angle below $90^\circ$ gives a positive inner product and hence
$h_{i,t}>0$, as illustrated in Figure~\ref{fig:zpd-alignment}.
Fisher cost still controls the acceptance magnitude.
Equation~\eqref{eq:zpd-controller-main} assigns zero weight to
nonpositive alignment.
\end{minipage}\hfill
\begin{minipage}[c]{0.22\columnwidth}
\centering
\includegraphics[width=\linewidth]{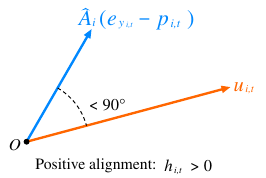}
\label{fig:zpd-alignment}
\end{minipage}
\par\smallskip
\noindent
\vspace{-2.2em}
\paragraph{Token-wise Localization.}
Despite the shared advantage $\widehat A_i$, prefix-dependent $p_{i,t}$ and
$u_{i,t}$ give token-specific weights. The effective weight $m_{i,t}w_{i,t}^{*}$
combines evidence eligibility with local acceptance, independently of restoration.
\vspace{-1em}
\paragraph{Cost-aware Scaling.}
A small logit step $\delta u_{i,t}$ incurs second-order KL movement
$\tfrac12\delta^2c_{i,t}$. Appendix~\ref{sec:app-zpd-proxies} derives this
Fisher approximation. We combine movement cost with a positive floor as
$\Lambda_{i,t}=\alpha_{\mathrm{cost}}c_{i,t}+\epsilon_{\mathrm{cost}}>0$, where
$\alpha_{\mathrm{cost}}\geq0$ scales cost and $\epsilon_{\mathrm{cost}}>0$
keeps the denominator positive. We then maximize the surrogate
$f_{i,t}(w)=h_{i,t}\log w+\Lambda_{i,t}\log(1-w)$ over $0<w<1$.
For positive benefit, its first term discourages zero acceptance, while
the second penalizes full acceptance. This surrogate balances the local
proxies without asserting an exact reward improvement.
For $h_{i,t}>0$, strict concavity and
$f'_{i,t}(w)=h_{i,t}/w-\Lambda_{i,t}/(1-w)=0$
give a unique maximizer, extended to zero at $h_{i,t}=0$:
\begin{equation}
\boxed{
w_{i,t}^{*}
=\sg\!\left[
\frac{h_{i,t}}{h_{i,t}+\Lambda_{i,t}}
\right]
=\sg\!\left[
\frac{h_{i,t}}
{h_{i,t}+\alpha_{\mathrm{cost}}c_{i,t}+\epsilon_{\mathrm{cost}}}
\right].}
\label{eq:zpd-controller-main}
\end{equation}
The stopped weight lies in $[0,1)$, increases with benefit at fixed cost,
and decreases with cost when both benefit and $\alpha_{\mathrm{cost}}$ are
positive. Otherwise it is nonincreasing. Stop-gradient blocks controller
backpropagation. Appendix~\ref{sec:app-zpd-surrogate} proves these properties
and the zero-benefit convention.

\subsection{Fix, Contrastive, and Fisher Evidence Contrast}
\label{sec:fec}
\begin{wrapfigure}{R}{0.5\columnwidth}    
\centering
\vspace{\dimexpr-\intextsep-5pt\relax}
\includegraphics[width=\linewidth]{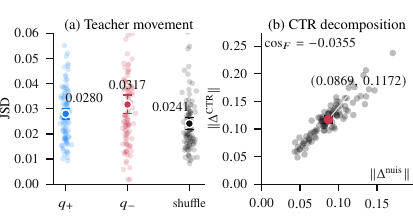}    
\captionsetup{        
width=\linewidth,        
font=small,        
skip=3pt,        
justification=raggedright,        
singlelinecheck=false    
}    
\caption{%        
Evidence-presence shifts motivate FEC.       
(a) divergence under correct, incorrect, and shuffled evidence;        
(b) task-contrast versus evidence-presence displacement norms.    
}    
\label{fig:fec-motivation}
\vspace{\dimexpr-\intextsep-\baselineskip\relax}
\end{wrapfigure}
Fisher Evidence Contrast (FEC) subtracts the task contrast's regularized
Fisher projection onto a measured evidence-presence direction.
Figure~\ref{fig:fec-motivation} motivates this separation: incorrect and
shuffled evidence also shift the Teacher, although this shift need not be
entirely task-irrelevant.
\vspace{-1em}
\paragraph{Evidence Contexts.}
The frozen Teacher branches $q^0$, $q^+$, and $q^-$ score the same sampled
prefix with no evidence, the ground-truth answer, and a wrong answer, respectively.
\vspace{-1em}

\paragraph{Endpoint Comparisons.} Fix~\citep{hubotter2026sdpo} uses
$\Delta^{\mathrm{Fix}}=q^+-q^0$.
Contrastive correction, denoted CTR, follows RLCSD~\citep{pan2026rlcsd}
with $\Delta^{\mathrm{CTR}}=q^+-q^-$.

\vspace{-1em}
\paragraph{FEC Refinement.} To attenuate the contrast along the measured
evidence-presence direction, we average the correct and incorrect Teacher
branches and compare this midpoint with the evidence-free branch. We then
measure the task contrast's component along this direction under the Fisher
metric with a positive stabilizer $\epsilon_{\mathrm{proj}}>0$:
\begin{equation}
\Delta^{\mathrm{nuis}}
= \tfrac12\left(q^+ + q^-\right) - q^0, \qquad
\alpha^{\FEC}
= \frac{
(\Delta^{\mathrm{CTR}})^\top F(p)\Delta^{\mathrm{nuis}}
}{
(\Delta^{\mathrm{nuis}})^\top F(p)\Delta^{\mathrm{nuis}}
+\epsilon_{\mathrm{proj}}
}.
\label{eq:main-fec-coefficient}
\end{equation}
The coefficient normalizes the Fisher inner product by the regularized
squared length of the presence proxy. Subtracting this component gives
\begin{equation}
\Delta^{\FEC}
=\Delta^{\mathrm{CTR}}-\alpha^{\FEC}\Delta^{\mathrm{nuis}}.
\label{eq:main-fec-direction}
\end{equation}
Regularization attenuates the residual's Fisher inner product with the
nuisance proxy without guaranteeing exact orthogonality.
We stop gradients through all of $\Delta^{\FEC}$, including its
Actor-dependent coefficient, and use it in the evidence loss and controller.
Table~\ref{tab:direction-constructions} summarizes the variants alongside
a geometric illustration of the endpoint comparisons and FEC refinement.
Appendix~\ref{sec:app-directions} derives the residual inner product and RKL construction.

\begin{table}[H]
\centering
\caption{FKL evidence-direction constructions and FEC geometry. Endpoint arrows point toward the subtracted branch. FEC removes the regularized nuisance component from CTR.}
\label{tab:direction-constructions}
\begin{minipage}[c]{0.52\textwidth}
\small
\hypersetup{hidelinks}
\setlength{\tabcolsep}{0pt}
\renewcommand{\arraystretch}{1.25}
\begin{tabular*}{\linewidth}{@{\extracolsep{\fill}}lcc@{}}
\toprule
Variant & Branches & FKL correction \\
\midrule
Fix~\citeyearpar{hubotter2026sdpo} & $q^0\rightarrow q^+$ & $\Delta^{\mathrm{Fix}}=q^+-q^0$ \\
CTR~\citeyearpar{pan2026rlcsd} & $q^-\rightarrow q^+$ & $\Delta^{\mathrm{CTR}}=q^+-q^-$ \\
FEC(Ours) & $q^0,q^-,q^+$ & $\Delta^{\mathrm{FEC}}=\Delta^{\mathrm{CTR}}-\alpha^{\mathrm{FEC}}\Delta^{\mathrm{nuis}}$ \\
\bottomrule
\end{tabular*}
\end{minipage}\hfill%
\begin{minipage}[c]{0.42\textwidth}
\centering
\includegraphics[width=\linewidth,trim=15bp 18bp 6bp 23bp,clip]{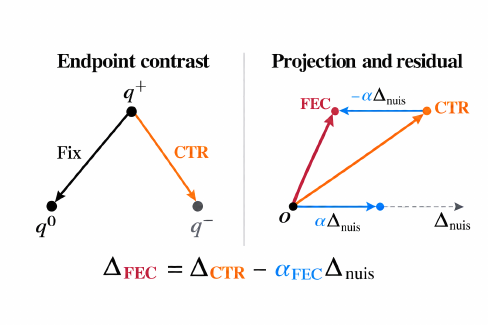}
\end{minipage}
% Offset the bottom float padding before the following subsection.
\par\vspace{-10pt}
\end{table}

\subsection{Two Actor-Update Paths}
\label{sec:two-paths}

The stopped controller supports two paths, evaluated separately.
Path~1 modulates the GRPO~\citep{shao2024deepseekmath} advantage, while
Path~2 scales the signed correction. For Path~1, the masked weight
$\widetilde w_{i,t}=m_{i,t}w_{i,t}^{*}$ combines evidence eligibility with
controller acceptance at each sampled token. For $\lambda_{\mathrm{adv}}\geq0$, the modulated advantage is
\begin{equation}
\widehat A'_{i,t}=\widehat A_i[1+\chi_{\mathrm{adv}}\lambda_{\mathrm{adv}}
\sg(\widetilde w_{i,t})].
\label{eq:advantage-modulation}
\end{equation}
The positive stopped multiplier preserves the advantage sign and clipping
decision, equals one for zero or ineligible acceptance, and contributes no
gradient. With switches $\chi_{\mathrm{OPSD}},\chi_{\mathrm{adv}}\in\{0,1\}$,
the joint objective is
\begin{equation}
\boxed{
\mathcal L_{\mathrm{total}}^{\FKL}
=\mathcal L_{\mathrm{GRPO}}(\widehat A')
+\lambda_{\mathrm{ref}}\mathcal L_{\mathrm{ref}}^{\FKL}
+\chi_{\mathrm{OPSD}}\lambda_{\mathrm{evi}}
\mathcal L_{\mathrm{OPSD}}^{\FKL}(w^{*}).}
\label{eq:verpo-objective}
\end{equation}

The outcome loss supplies the task signal, while $\lambda_{\mathrm{ref}}$
and $\lambda_{\mathrm{evi}}$ scale reference restoration and accepted correction.
Path~1 uses $(\chi_{\mathrm{adv}},\chi_{\mathrm{OPSD}})=(1,0)$ and Path~2
uses $(0,1)$. Algorithm~\ref{alg:verpo-zpd-short} specifies masking,
normalization, and the joint update.

\begin{algorithm}[H]
\caption{\method{} Actor Update under FKL Geometry}
\label{alg:verpo-zpd-short}
\small
\begin{algorithmic}[1]
\REQUIRE Actor $\pi_\theta$, EMA Teacher $\pi_{\bar\theta}$,
  data $\mathcal D$, variant $s$,
  switches $\chi_{\mathrm{OPSD}},\chi_{\mathrm{adv}}$

\REPEAT
  \STATE Freeze $\theta_{\mathrm{old}}\leftarrow\theta$;
    sample response groups from $\pi_{\theta_{\mathrm{old}}}$
    for prompts in $\mathcal D$.

  \STATE Obtain verifier rewards $R_i$; set
    $\widehat A_i\leftarrow
      R_i-|\mathcal G(i)|^{-1}\sum_{j\in\mathcal G(i)}R_j$.
    \VERPOcomment{Outcome anchor}

  \STATE Score Actor prefixes as $p_{i,t}$; replay them with
    stopped Teacher branches
    $q^{\mathrm{ref}}_{i,t},q^0_{i,t},q^+_{i,t},q^-_{i,t}$.

  \STATE Select \VERPOhl{verpoRed}{\Delta_{i,t}}:
    $q^+_{i,t}-q^0_{i,t}$ (Fix),
    $q^+_{i,t}-q^-_{i,t}$ (CTR),
    or Eq.~\eqref{eq:main-fec-direction} (FEC).
    \VERPOcomment[verpoRed]{Signed correction}

  \STATE Set $u_{i,t}\leftarrow\sg[\Delta_{i,t}]$;
    compute $h_{i,t},c_{i,t}$ via Eq.~\eqref{eq:zpd-proxies-main}.
    \VERPOcomment[verpoBlue]{Alignment and cost}

  \STATE Set
    \VERPOhl{verpoOrange}{
      w^*_{i,t}\leftarrow
      \sg[h_{i,t}/(h_{i,t}
        +\alpha_{\mathrm{cost}}c_{i,t}
        +\epsilon_{\mathrm{cost}})]
    }.
    \VERPOcomment[verpoOrange]{Token-wise ZPD gate}

  \STATE Form masks $a_{i,t}$ (valid) and $m_{i,t}\leq a_{i,t}$ (evidence);
    set $\widetilde w_{i,t}\leftarrow m_{i,t}w^*_{i,t}$.

  \STATE Set $N\leftarrow\sum_{i,t}a_{i,t}$ and
    $Z_{\mathrm{evi}}\leftarrow\max(1,\sum_{i,t}a_{i,t}m_{i,t})$.

  \STATE Form $\widehat A'_{i,t}$ via Eq.~\eqref{eq:advantage-modulation};
    compute normalized losses via Eqs.~\eqref{eq:main-fkl-reference-loss}
    and~\eqref{eq:main-fkl-opsd-loss}.
    \VERPOcomment[verpoBlue]{Two-path update}

  \STATE Update
    $\theta\leftarrow\theta-\eta_\theta
      \nabla_\theta\mathcal L_{\mathrm{total}}^{\FKL}$
    via Eq.~\eqref{eq:verpo-objective};
    update the EMA Teacher via Eq.~\eqref{eq:app-ema-teacher}.

\UNTIL{stopping criterion is reached}
\RETURN $\pi_\theta$
\end{algorithmic}
\end{algorithm}

% Keep result floats close while retaining the template text dimensions.
\begingroup
\setlength{\floatsep}{6pt plus 1pt minus 1pt}
\setlength{\textfloatsep}{8pt plus 1pt minus 1pt}
\setlength{\intextsep}{8pt plus 1pt minus 1pt}
\section{Experiments}
\label{sec:experiments}

\subsection{Experimental Setup}
We use instruction-tuned Qwen3-4B, Qwen3-8B~\citep{yang2025qwen3},
and Llama-3.2-1B-Instruct~\citep{meta2024llama32} as backbones.
Following the evaluation setup of SDPO~\citep{hubotter2026sdpo},
we consider five benchmarks: Chemistry, Physics, Biology, Materials,
and Tool Use. The four science benchmarks use the reasoning subsets
of SciKnowEval~\citep{feng2024sciknoweval} to evaluate undergraduate-level
scientific reasoning, while Tool Use uses
ToolAlpaca~\citep{tang2023toolalpaca} to evaluate tool-call generation
given a user request and tool specifications.
We compare against GRPO~\citep{shao2024deepseekmath}, which learns
from group-relative outcome rewards;
SDPO~\citep{hubotter2026sdpo}, which obtains dense token supervision
from a feedback-conditioned teacher;
SRPO~\citep{li2026srpo}, which routes successful samples to the GRPO
objective and unsuccessful samples to the SDPO objective;
RLSD~\citep{yang2026rlsd}, which uses verifier rewards to modulate
self-distillation; and RLCSD~\citep{pan2026rlcsd}, which contrasts
teachers conditioned on correct and incorrect hints.
We evaluate two \method{} variants: VERPO-LW applies the ZPD weights
to the evidence loss, while VERPO-AM uses them to modulate
the GRPO advantages.
Table~\ref{tab:main-results} reports each task's best-checkpoint avg@16
and their unweighted mean as Average.
The test split also supplies the logged validation scores used for selection.
Appendix~\ref{sec:app-details} clarifies this protocol and its limitations.

% \section{Experiment Results}

\subsection{Main Results}

\begin{table}[h]
\centering
\small
\caption{Best-checkpoint avg@16 results. Base scores appear in the backbone rows.
In the Path column, 1 denotes advantage modulation, which rescales token
advantages, and 2 denotes weighted correction, which scales token-level
evidence corrections.
RH denotes reward hacking, blank entries denote unavailable scores, and
$\dagger$ marks collapse or an average containing one.
SDPO Physics on Llama-3.2-1B reports its pre-collapse peak.}
\label{tab:main-results}
\renewcommand{\arraystretch}{1.04}
\setlength{\tabcolsep}{3.4pt}
\begin{tabular*}{\textwidth}{@{\extracolsep{\fill}}clcccccc@{}}
\toprule
Path & Method & Biology & Chemistry & Materials & Physics & Tool Use & Average \\
\midrule

\multicolumn{2}{@{}l}{\textbf{Qwen3-8B}}
& 28.12 & 40.86 & 58.44 & 60.85 & 58.27 & 49.31 \\
\cmidrule(lr){1-8}
1 & RLSD~\citep{yang2026rlsd}
& 53.50 & 69.82 & 76.06 & 66.17 & 64.06 & 65.92 \\
 & RLCSD~\citep{pan2026rlcsd}
& 38.62 & 62.94 & 66.88 & 51.01 & 54.50 & 54.79 \\
 & VERPO-AM~(Ours)
& 59.13 & \underline{78.46} & \underline{76.79}
& \textbf{69.09} & \textbf{69.42} & \underline{70.58} \\

\cmidrule(lr){1-8}
2 & GRPO~\citep{shao2024deepseekmath}
& \underline{61.75} & 70.38 & 76.66 & 66.87 & 67.18 & 68.57 \\
 & SDPO~\citep{hubotter2026sdpo}
& 60.75 & 77.94 & 74.93 & 64.06 & 67.09 & 68.95 \\
 & SRPO~\citep{li2026srpo}
& 60.00 & 73.51 & 73.27 & 68.51 & 66.36 & 68.33 \\
 & VERPO-LW~(Ours)
& \textbf{62.37} & \textbf{80.26} & \textbf{77.62}
& \underline{68.85} & \underline{68.15} & \textbf{71.44} \\

\midrule

\multicolumn{2}{@{}l}{\textbf{Qwen3-4B}} & 33.25 & 44.37 & 60.90 & 61.40 & 58.91 & 51.77\\
\cmidrule(lr){1-8}

1 & RLSD~\citep{yang2026rlsd}
& 51.50 & 69.13 & 77.65 & 68.51 & 60.75 & 65.51 \\
 & RLCSD~\citep{pan2026rlcsd}
& 47.25 & 60.13 & 67.15 & 50.93 & 58.91 & 56.87 \\
 & VERPO-AM~(Ours)
& 55.37 & \textbf{72.85} & 75.44
& \textbf{70.78} & 59.09 & 66.71 \\

\cmidrule(lr){1-8}
2 & GRPO~\citep{shao2024deepseekmath}
& 52.25 & 70.00 & \textbf{80.18} & 67.42 & 60.56 & 66.08 \\
 & SDPO~\citep{hubotter2026sdpo}
& 47.87 & 69.79 & 72.40 & 62.89 & 57.72 & 62.13 \\
 & SRPO~\citep{li2026srpo}
& \underline{60.12} & \underline{72.47} & 78.78
& 67.89 & \textbf{62.04} & \underline{68.26} \\
 & VERPO-LW~(Ours)
& \textbf{62.00} & 69.17 & \underline{79.72}
& \underline{70.55} & \underline{61.40} & \textbf{68.57} \\

\midrule

\multicolumn{2}{@{}l}{\textbf{Llama-3.2-1B}} & 10.62 & 7.41 & 5.59 & 5.55 & 1.47 & 6.13 \\

\cmidrule(lr){1-8}
1 & RLSD~\citep{yang2026rlsd}
& RH & RH & RH & RH & RH & RH \\
 & RLCSD~\citep{pan2026rlcsd}
& 15.87 & 24.97$^\dagger$ & 30.71
& 20.85 & 33.45 & 25.17$^\dagger$ \\
 & VERPO-AM~(Ours)
& \underline{55.50} & 63.69 & \textbf{55.06}
& 50.70 & \textbf{51.01} & \underline{55.19} \\

\cmidrule(lr){1-8}
2 & GRPO~\citep{shao2024deepseekmath}
& 42.38 & 52.27 & 42.93 & \underline{52.47} & 47.51 & 47.51 \\
 & SDPO~\citep{hubotter2026sdpo}
& 31.50 & 30.47 & 29.38 & 40.15$^\dagger$ & 43.85 & 35.07$^\dagger$ \\
 & SRPO~\citep{li2026srpo}
& 31.87 & \textbf{67.11} & 31.91 & 44.45 & 42.15 & 43.50 \\
 & VERPO-LW~(Ours)
& \textbf{60.15} & \underline{66.57} & \underline{52.26}
& \textbf{54.22} & \underline{49.63} & \textbf{56.57} \\

\bottomrule
\end{tabular*}
\vspace{-1em}
\end{table}

Table~\ref{tab:main-results} shows that VERPO-LW achieves the highest
five-task best-checkpoint average on every backbone among methods with complete and
noncollapsed results.
On Llama-3.2-1B, RLSD~\citep{yang2026rlsd} exhibits reward hacking,
while SDPO~\citep{hubotter2026sdpo} and RLCSD~\citep{pan2026rlcsd}
include collapsed runs and are excluded from average ranking.
The SDPO Physics score is a pre-collapse peak, not its final performance.
Its descriptive five-task average retains this peak and carries the same collapse marker.
The preferred update path varies by task.
On Qwen3-4B, VERPO-LW achieves an average of 68.57, exceeding the strongest
baseline by 0.31 points.
On Llama-3.2-1B, it reaches 56.57, improving over the strongest baseline
with complete and noncollapsed results by 9.06 points.
On Qwen3-8B, VERPO-LW exceeds the strongest baseline average by 2.49 points.
The compute-matched comparison in
Appendix~\ref{sec:compute-matched-performance} also favors VERPO on
Llama-3.2-1B under a matched GPU budget.
\vspace{-1em}
\subsection{Ablation Study}
\label{sec:ablation}

\paragraph{Component Ablations.}
Table~\ref{tab:component-ablation} compares Teacher update rules, evidence
directions, and KL geometries on Llama-3.2-1B. The more responsive EMA Teacher
performs best among the tested update rules, while FEC consistently outperforms
CTR~\citep{pan2026rlcsd} across all tasks. The Fixed direction exhibits reward
hacking on Biology and Chemistry, with the remaining task results unavailable.
FKL provides complete results across all tasks, whereas the evaluated RKL runs
collapse on Biology and Chemistry and are unavailable elsewhere. 
% \vspace{-2em}
\begin{table}[t]
\centering
\caption{(a) Llama-3.2-1B component. Teacher, direction, and geometry
comparisons.
(b) Chemistry evidence-loss scale. avg@16 is the step-200 score.
Late reports the mean and sample standard deviation over 11 evaluations at steps 150--200.
(c) Chemistry uniform weights. L1B/Q4B/Q8B denote Llama-3.2-1B/Qwen3-4B/Qwen3-8B.
RH denotes reward hacking, -- unavailable scores, and $\dagger$ collapse.
Marked scores in (c) are pre-collapse peaks.}
\label{tab:ablation-summary}
\captionsetup[subtable]{labelformat=simple,font=small,labelfont=bf,textfont=bf,
justification=raggedright,singlelinecheck=false,skip=4pt}
\renewcommand{\thesubtable}{(\alph{subtable})}
\begin{subtable}[t]{0.725\textwidth}
\vspace{0pt}
\caption{Component ablations}
\label{tab:component-ablation}
\footnotesize
\renewcommand{\arraystretch}{1.04}
\setlength{\tabcolsep}{0pt}
% Give every score column the same width and center its heading and values.
\begin{tabular}{@{}p{51pt}*{6}{>{\centering\arraybackslash}p{\dimexpr(\linewidth-51pt)/6\relax}}@{}}
\toprule
Setting & Biology & Chemistry & Materials & Physics & Tool Use & Average \\
\midrule
\multicolumn{7}{@{}l}{\textit{Teacher source}} \\
\cmidrule(lr){1-7}
Fixed Teacher & 28.50 & 46.93 & 20.07 & 23.82 & 41.62 & 32.19 \\
EMA-0.99 & 49.75 & 62.41 & 38.03 & 46.32 & 48.80 & 49.06 \\
EMA-0.95 & 60.15 & 66.57 & 52.26 & 54.22 & 49.63 & 56.57 \\
\midrule
\multicolumn{7}{@{}l}{\textit{Evidence direction}} \\
\cmidrule(lr){1-7}
Fixed~\citeyearpar{hubotter2026sdpo} & RH & RH & -- & -- & -- & -- \\
CTR~\citeyearpar{pan2026rlcsd} & 57.75 & 65.86 & 44.89 & 52.09 & 47.19 & 53.56 \\
FEC & 60.15 & 66.57 & 52.26 & 54.22 & 49.63 & 56.57 \\
\midrule
\multicolumn{7}{@{}l}{\textit{KL geometry}} \\
\cmidrule(lr){1-7}
FKL & 60.15 & 66.57 & 52.26 & 54.22 & 49.63 & 56.57 \\
RKL & 20.87\rlap{$^\dagger$} & 31.10\rlap{$^\dagger$} & -- & -- & -- & -- \\
\bottomrule
\end{tabular}
\end{subtable}
\hfill
\begin{minipage}[t]{0.26\textwidth}
\raggedright
\setlength{\parindent}{0pt}
\setlength{\parskip}{0pt}
\vspace{0pt}
% Verified against the validation_curves.csv and metrics.json exports in
% PGR-Probe/reports/llama3_2_1b_chemistry_lambda_evi_sensitivity/.
% Late variation is across checkpoints, not across independent seeds.
\begin{subtable}[t]{\linewidth}
\vspace{0pt}
\caption{Evidence-loss scale}
\label{tab:lambda-evi-ablation}
\footnotesize
\setlength{\tabcolsep}{2pt}
\renewcommand{\arraystretch}{1.04}
\begin{tabular*}{\linewidth}{@{\extracolsep{\fill}}ccc@{}}
\toprule
$\lambda_{\mathrm{evi}}$ & avg@16 & Late \\
\midrule
0.2 & 65.09 & $63.35\pm1.41$ \\
0.4 & 65.00 & $63.51\pm0.96$ \\
0.6 & 61.28 & $62.78\pm1.92$ \\
0.8 & 62.11 & $60.63\pm1.29$ \\
1.0$^*$ & 66.58 & $65.52\pm1.09$ \\
\bottomrule
\end{tabular*}
\end{subtable}

% Align the bottom rules of panels (a) and (c) at the current row spacing.
\par\vspace{-3.123pt}
% Values supplied by the author in the annotated Chemistry ablation table.
% A dagger marks an author-identified collapse and a pre-collapse peak.
\begin{subtable}[t]{\linewidth}
\vspace{0pt}
\caption{Uniform weights}
\label{tab:uniform-weight-ablation}
\footnotesize
\setlength{\tabcolsep}{1pt}
\renewcommand{\arraystretch}{1.04}
\begin{tabular*}{\linewidth}{@{\extracolsep{\fill}}crrr@{}}
\toprule
$w$ & L1B & Q4B & Q8B \\
\midrule
0.25 & 53.48 & 71.99 & 73.30 \\
% Keep markers outside the numeric alignment width.
0.50 & 20.88\rlap{$^\dagger$} & 68.89 & 63.06 \\
0.75 & 30.47\rlap{$^\dagger$} & 46.57\rlap{$^\dagger$} & 63.51 \\
1.00 & 20.53\rlap{$^\dagger$} & 46.16\rlap{$^\dagger$} & 37.76 \\
\bottomrule
\end{tabular*}
\end{subtable}

\end{minipage}
\end{table}

\vspace{-1em}
\paragraph{Evidence-Loss Scale.}
We use $\lambda_{\mathrm{evi}}=1.0$ as the default.
Its configuration has the highest step-200 score in
Table~\ref{tab:lambda-evi-ablation}. All five settings use all-trajectory
correction, but code revision and planned horizon differ for 1.0.
This comparison does not isolate the coefficient effect or establish an
optimal value. Appendix~\ref{sec:hyperparameter_sensitivity} details the protocol.

\vspace{-1em}
\paragraph{Uniform Evidence Weights.}
\label{sec:uniform-weight-ablation}
To test the role of token-wise acceptance, we replace ZPD with a constant
$w$ in VERPO-LW while retaining the remaining settings within each backbone.
Table~\ref{tab:uniform-weight-ablation} shows two trends on Chemistry.
At the same uniform weight, smaller evaluated backbones are more prone to
collapse. Within each backbone, increasing the uniform weight generally
worsens validation performance, although the trend is not strictly monotonic.
These observations support the instability concern in the introduction:
indiscriminate evidence correction can destabilize training, with smaller
backbones showing greater vulnerability in the tested settings.
Marked entries report pre-collapse peaks rather than final scores.

\vspace{-1em}
\subsection{Further Analysis}

\paragraph{Training Dynamics.}
\label{further:training_dynamic}

To assess how performance develops during training,
Figure~\ref{fig:training-dynamics-physics} tracks the 200-step Physics runs
on Llama-3.2-1B. VERPO finishes with higher validation accuracy than
GRPO~\citep{shao2024deepseekmath}, although their accuracy curves cross
during training. The SDPO~\citep{hubotter2026sdpo} run reaches zero
validation accuracy at the displayed checkpoints from step 160 onward,
after the pre-collapse peak reported in Table~\ref{tab:main-results}. VERPO retains longer responses late in training
than the plotted baselines, while its Actor entropy remains low.

\begin{figure}[!htb]
   \centering
  \includegraphics[width=\textwidth]{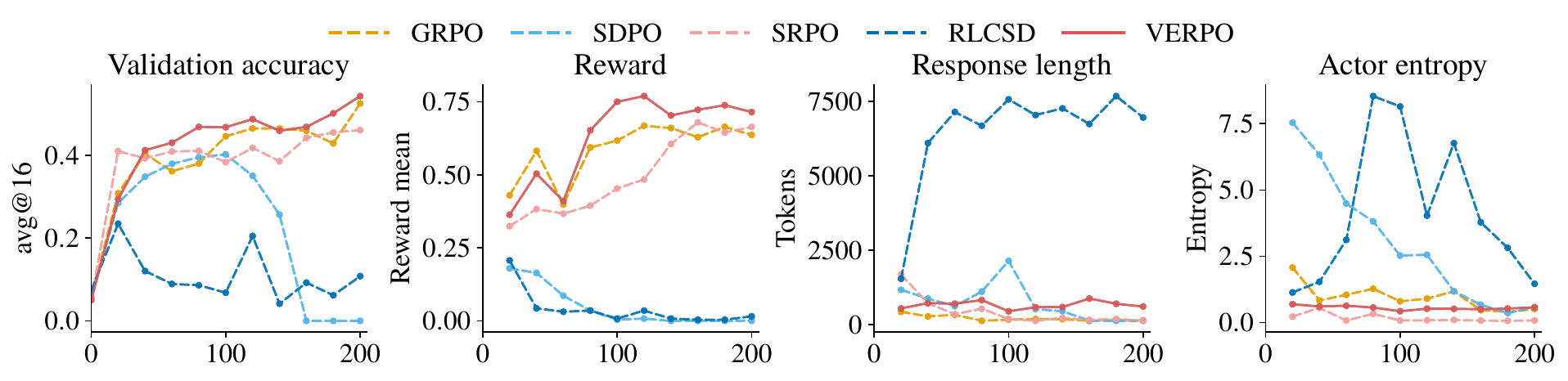}
   \caption{Physics training on Llama-3.2-1B for 200 steps: validation avg@16
$\uparrow$, reward, response length, and Actor entropy.
Methods are GRPO~\citep{shao2024deepseekmath}, SDPO~\citep{hubotter2026sdpo},
SRPO~\citep{li2026srpo}, RLCSD~\citep{pan2026rlcsd}, and VERPO.}
    \label{fig:training-dynamics-physics}
\end{figure}

\begin{figure}[!htb]
    \centering
\includegraphics[width=\textwidth]{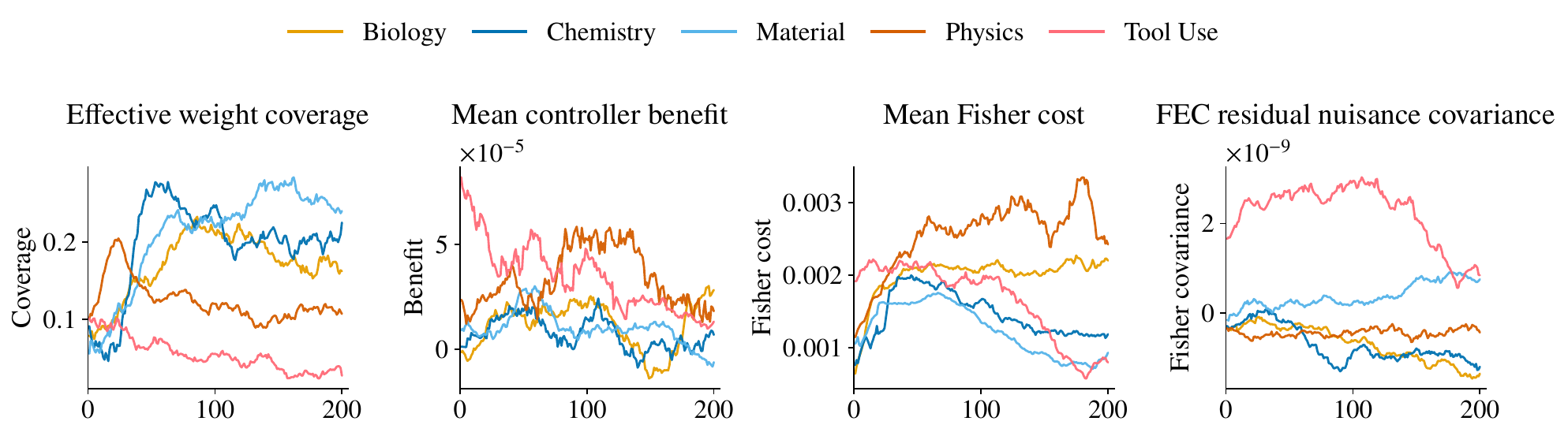}
\caption{Llama-3.2-1B diagnostics across five tasks: effective-weight coverage
($w_{i,t}>10^{-3}$), signed benefit, Fisher cost, and residual--nuisance covariance.
Appendix~\ref{sec:app-diagnostics} defines the metrics.
Near-zero mean covariance does not imply token-wise orthogonality.}
    \label{fig:fec-diagnostics}
\end{figure}

\begin{figure}[!htb]
    \centering
    \includegraphics[width=\textwidth]{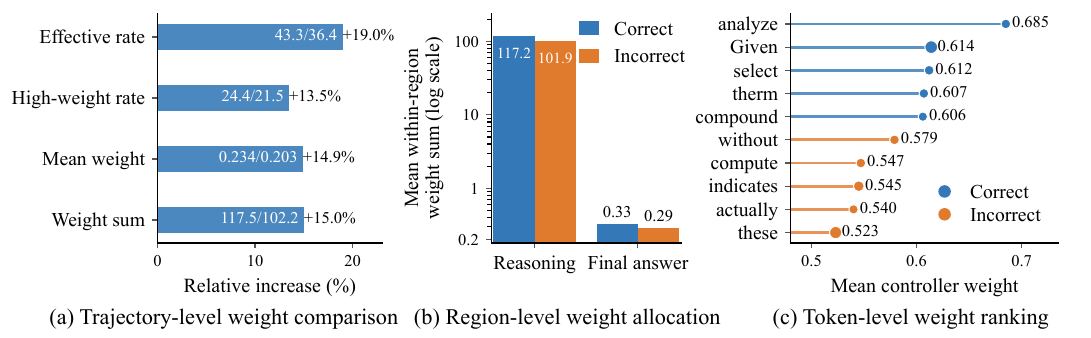}
\caption{FEC and ZPD acceptance in matched Materials responses:
(a) trajectory classes, (b) regional weight sums on a log scale, and
(c) token-type mean weights with frequency-scaled markers.
Effective/high weights satisfy $w>10^{-3}$/$w\geq0.5$.
Token types occur at least 20 times.}
    \label{fig:fec-token-analysis-summary}
\end{figure}

\vspace{-1em}
\paragraph{FEC Diagnostics.}
\label{further:fec_dianose}

Figure~\ref{fig:fec-diagnostics} reports task-wise effective-weight coverage,
signed pre-clamp benefit, Fisher cost, and residual nuisance covariance. Near
the final displayed step, Materials shows broader coverage than Tool Use, but
this reflects the extent of evidence acceptance rather than relative task
benefit, and cannot be reconstructed from aggregate benefit and cost alone.
The near-zero mean residual covariance supports the regularized projection
only in aggregate and may conceal token-level cancellation. Definitions and
the individual-direction identity are given in
Appendices~\ref{sec:app-diagnostics} and~\ref{sec:app-directions}.

\vspace{-1em}
\paragraph{Token-Level FEC Weight Analysis.}
\label{further:token_level_fec_weight}
Figure~\ref{fig:fec-token-analysis-summary} examines where FEC and ZPD accept
evidence in matched Materials responses. Correct rollouts show broader
acceptance, while both trajectory classes concentrate most weight in
intermediate reasoning rather than final-answer tokens, reflecting both region
length and higher per-token weights. Token-type rankings illustrate lexical
allocation patterns but characterize acceptance rather than causal error
contributions. Appendix~\ref{sec:token-weight-analysis} reports the complete
statistics, and Appendix~\ref{sec:app-case-study} provides a rollout-level
example.

\section{Conclusion}

In this work, we introduce VERPO, a framework combining verifiable outcome rewards with dense evidence-conditioned supervision without assuming equal reliability of Teacher proposals. VERPO separates evidence-free reference restoration from signed evidence-induced correction, uses FEC to remove the measured evidence-presence component, and applies a token-wise ZPD controller to selectively accept corrections based on alignment with the local GRPO direction and associated Fisher movement costs. Across five benchmarks and three model backbones, VERPO achieves the best average performance within each model block. Ablation studies show that stability depends on the Teacher update rule, evidence construction, and KL geometry, while fine-grained analysis finds accepted corrections concentrated mainly on intermediate reasoning rather than final-answer tokens.

\clearpage
\endgroup
\endgroup
\bibliography{references}
\bibliographystyle{iclr2027_conference}

\section*{AI Use Statement}

We used generative AI tools solely to assist with manuscript drafting and English-language editing. The authors critically reviewed and revised all AI-assisted content and made all final decisions regarding the content and presentation of the manuscript. The authors take full responsibility for the final version of this work.

\appendix
\numberwithin{equation}{section}

\clearpage
\section*{Appendix Contents}
\addcontentsline{toc}{section}{Appendix Contents}

\begingroup
\setlength{\parskip}{0.35em}

\newcommand{\appcontentsline}[2]{%
  \noindent\hyperref[#1]{\textbf{\ref*{#1}}\quad #2}%
  \dotfill\hyperref[#1]{\pageref*{#1}}\par}
  
\newcommand{\appcontentssubline}[2]{%
  \noindent\hspace*{1.5em}\hyperref[#1]{\ref*{#1}\quad #2}%
  \dotfill\hyperref[#1]{\pageref*{#1}}\par}

\appcontentsline{sec:app-related-work}{Extended Related Work}

\medskip
\appcontentsline{sec:notation}{Notation}

\medskip
\appcontentsline{sec:algorithm}{Algorithm}

\medskip
\appcontentsline{sec:preliminaries}{Preliminaries}
\appcontentssubline{sec:app-notation}{Autoregressive Policy}
\appcontentssubline{sec:app-grpo-prelim}{RLVR and Group-Relative Policy Optimization}
\appcontentssubline{sec:app-teacher-replay}{Feedback-Conditioned Teacher Replay}
\appcontentssubline{sec:app-local-geometry}{Divergence and Local Information Geometry}

\medskip
\appcontentsline{sec:theoretical-analysis}{Theoretical Analysis}
\appcontentssubline{sec:app-teacher-paths}{Teacher Paths}
\appcontentssubline{sec:app-fkl}{Exact Forward-KL Decomposition}
\appcontentssubline{sec:app-rkl}{Reverse-KL Counterpart}
\appcontentssubline{sec:app-directions}{Fixed, Contrastive, and Fisher Evidence Contrast}
\appcontentssubline{sec:app-zpd-ideal}{Ideal Token-Wise Controller}
\appcontentssubline{sec:app-zpd-proxies}{Benefit and Local-KL Cost Proxies}
\appcontentssubline{sec:app-zpd-surrogate}{Surrogate Acceptance Optimization and Properties}
\appcontentssubline{sec:app-advantage-modulation}{Multiplicative Advantage Modulation}
\appcontentssubline{sec:app-complete-objective}{Complete Objective and Exact Logit Gradients}

\medskip

\appcontentsline{sec:app-additional-results}{Additional Results}
\appcontentssubline{sec:app-joint-paths}{Joint Loss Weighting and Advantage Modulation}
\appcontentssubline{sec:app-correction-scope}{Trajectory-Level Correction Scope}
\appcontentssubline{sec:token-weight-analysis}{Token-Level FEC and ZPD Weight Analysis}
\appcontentssubline{sec:hyperparameter_sensitivity}{Hyperparameter Sensitivity}
\appcontentssubline{sec:app-topk-support}{Probability Mass Retention under Top-K Truncation}

\medskip

\appcontentsline{sec:app-diagnostics}{Diagnostic Metrics}
\appcontentssubline{sec:diagnostic-zpd}{ZPD Effective Weight Coverage}
\appcontentssubline{sec:diagnostic-benefit-cost}{Mean Signed Benefit and Fisher Cost}
\appcontentssubline{sec:diagnostic-fec}{FEC Residual Nuisance Fisher Covariance}

\medskip

\appcontentsline{sec:app-details}{Experimental Details}
\appcontentssubline{sec:app-datasets}{Datasets and Data Processing}
\appcontentssubline{sec:app-technical-setup}{Models, Baselines, and Runtime}
\appcontentssubline{sec:app-shared-setup}{Shared Training Setup}
\appcontentssubline{sec:app-method-hyperparameters}{Method-Specific Hyperparameters}
\appcontentssubline{sec:app-prompts}{Prompt Templates}

\medskip

\appcontentsline{sec:computational-analysis}{Computational Analysis}
\appcontentssubline{sec:per-step-cost}{Per-Step Cost}
\appcontentssubline{sec:compute-matched-performance}{Compute-Matched Performance}

\medskip

\appcontentsline{sec:app-case-study}{Case Study}
\endgroup

\clearpage

\section{Extended Related Work}
\label{sec:app-related-work}

This appendix expands the discussion in Section~\ref{sec:related-work},
covering outcome-based credit assignment, self-distillation, and selective
privileged supervision in greater detail.

\paragraph{RLVR and Fine-grained Credit Assignment.}
Reinforcement learning with verifiable rewards (RLVR) optimizes language models
using automatically checkable outcome rewards. Critic-free methods such as
GRPO~\citep{shao2024deepseekmath} replace traditional actor--critic value
estimates with group-relative advantages computed from multiple on-policy
responses. DeepSeek-R1~\citep{deepseekai2025deepseekr1} and
DAPO~\citep{yu2025dapo} use related group-relative updates for scalable RLVR.
These sequence-level advantages are shared across response tokens, obscuring
which local decisions caused the outcome. Process supervision obtains
finer-grained signals through
human-annotated step verification in Let's Verify Step by Step~\citep{lightman2023lets},
automatically constructed labels in Math-Shepherd~\citep{wang2024mathshepherd},
and outcome-derived progress estimates in Rewarding Progress~\citep{setlur2025rewarding}.
Such approaches require human annotations or automatically generated process
targets together with a learned verifier, whose local preferences need not
match the verifiable sequence-level objective. VERPO retains the outcome-derived
GRPO advantage as the task-level anchor and uses privileged Teacher replay to
propose token-level corrections, accepting them only when they align with the
local reward direction at justified Fisher movement cost.

\paragraph{On-policy Distillation and Self-distillation.}
Knowledge distillation transfers a teacher's predictive distribution to a
student~\citep{hinton2015distilling}. On-policy distillation reduces off-policy
distribution shift by supervising student-generated trajectories with an
external teacher~\citep{agarwal2023onpolicy}. MiniLLM develops a related
distribution-matching formulation for language-model distillation~\citep{gu2023minillm}.
OPSD removes the external teacher by conditioning the same model on privileged
solutions~\citep{zhao2026opsd}, and SDPO constructs a feedback-conditioned
self-teacher for dense token-level supervision~\citep{hubotter2026sdpo}.
SRPO anchors self-distillation to verifiable outcomes through sample-level
routing~\citep{li2026srpo}. RLSD modulates the magnitude of self-distillation
with a reward-directed signal~\citep{yang2026rlsd}, and EviSD applies analogous
reward-anchored modulation to evidence-conditioned search actions~\citep{xie2026evisd}.
RLCSD uses contrastive privileged contexts to shape its self-distillation signal
~\citep{pan2026rlcsd}. These methods do not jointly separate evidence-free
reference retention from signed evidence-induced change, and they do not decide
whether an individual correction is worth applying at its local policy cost.
VERPO addresses this control gap with a stopped token-wise ZPD controller that
accepts and scales each correction according to its alignment with the local
GRPO direction and its Fisher movement cost.

\paragraph{Selective and Contrastive Privileged Supervision.}
Empirical studies show that privileged supervision is not uniformly reliable
across rollouts or tokens~\citep{armandpour2026unmasking,pibias2026break,
lazaridis2026edgeopd,zhao2026rosd,kaur2026rethinking,jiang2026saopd}.
Teacher guidance is better aligned on incorrect than on
already-correct trajectories~\citep{armandpour2026unmasking}, while privileged
conditioning can bias the Teacher toward a particular reference trajectory
rather than toward correctness in general~\citep{pibias2026break}, induce
behavior unrelated to the intended correction~\citep{lazaridis2026edgeopd},
overwrite valid reasoning prefixes
\citep{zhao2026rosd}, suppress verification and backtracking in thinking models
\citep{kaur2026rethinking}, or amplify weakly input-grounded signals
\citep{jiang2026saopd}. RLCSD addresses privilege-induced style drift by
contrasting self-teachers conditioned on correct and incorrect hints~\citep{pan2026rlcsd}. CEPO similarly exploits correct-wrong evidence for
contrastive token credit~\citep{heakl2026cepo}, while OCSD uses structurally
matched privileged and ablated replay views to remove replay-scaffold effects
in agentic RL~\citep{yang2026ocsd}. These contrastive constructions reduce
shared effects between privileged branches, but a Teacher difference can still
contain a component caused by the mere presence of privileged evidence. Such a
component can leak evidence-specific style, confidence, or formatting into the
Student without improving task success. VERPO's Fisher Evidence Contrast (FEC)
estimates this evidence-presence direction and projects it out of the task
contrast under the local Fisher metric, reducing privileged-context leakage
before the resulting correction is applied.

% Keep the heading, introduction and table at their natural vertical spacing.
% The appendix template otherwise stretches these gaps to fill this page.
\par\noindent\begin{minipage}{\textwidth}
\section{Notation}
\label{sec:notation}

Table~\ref{tab:theory-notation} aligns the notation in Section~\ref{sec:method},
Algorithm~\ref{alg:verpo-zpd} and Algorithm~\ref{alg:verpo-zpd-short} , and the derivations.
Symbols are grouped by their operational role. Algebraic dummy variables
are defined locally.

\begin{table}[H]
\centering
\fontsize{9}{10.5}\selectfont
\caption{Notation used throughout the method and appendices.}
\label{tab:theory-notation}
\renewcommand{\arraystretch}{1.0}
\begin{tabularx}{\textwidth}{@{}lY@{}}
\toprule
Symbol & Definition \\
\midrule
\multicolumn{2}{@{}l@{}}{\textbf{Data, policies, and rollout quantities}} \\

\midrule

$x_i,y_i,y_{i,<t},t,v,\mathcal V$ & Prompt, sampled response, response prefix, token position, vocabulary item, and vocabulary. \\
$\mathcal G(i)$ & Rollout group containing responses for the same prompt as $i$. \\
$R_i,\widehat A_i$ & Scalar outcome reward and group-relative sequence advantage. \\
$\theta,\theta_{\mathrm{old}},\bar\theta$ & Current Actor, frozen behavior Actor, and exponential-moving-average Teacher parameters. \\
$p_{i,t},p_{i,t}^{\mathrm{old}},z_{i,t}$ & Current/behavior token distributions and current Actor logits at prefix $y_{i,<t}$. \\
$\rho_{i,t}(\theta)$ & Token-level importance ratio in GRPO~\citep{shao2024deepseekmath}. \\
\midrule
\multicolumn{2}{@{}l@{}}{\textbf{Teacher branches and evidence paths}} \\

\midrule

$q_{i,t}^{(c)},c$ & Frozen Teacher distribution and its replay/context condition. \\
$q_{i,t}^{\mathrm{ref}}$ & Evidence-free reference Teacher used for policy restoration. \\
$q_{i,t}^{0},q_{i,t}^{e}$ & Teacher distributions without and with task evidence. \\
$q_{i,t}^{+},q_{i,t}^{-}$ & Correct-evidence Teacher and incorrect-evidence Teachers. \\
$q_{b,i,t},q_{e,i,t}$ & Base and target endpoints of a selected evidence comparison. \\
$q_{w,t}^{\FKL},q_{w,t}^{\RKL},Z_t(w)$ & Arithmetic/geometric Teacher path and its RKL partition function. \\
$\Delta_{i,t},r_{i,t}$ & Probability displacement $q_{e,i,t}-q_{b,i,t}$ and log-density contrast $\log q_{e,i,t}-\log q_{b,i,t}$. \\
$F(p_{i,t}),G_{i,t}$ & Categorical Fisher matrix $\diag(p_{i,t})-p_{i,t}p_{i,t}^{\top}$ and logit Jacobian $\partial z_{i,t}/\partial\theta$. \\
$u_{i,t}$ & Evidence logit direction: $\Delta_{i,t}$ for FKL and $F(p_{i,t})r_{i,t}$ for RKL. \\
\midrule
\multicolumn{2}{@{}l@{}}{\textbf{Direction constructions and FEC}} \\

\midrule

$\Delta_t^{\mathrm{task}},\Delta_t^{\mathrm{nuis}},\Delta_t^{\FEC}$ & Zero-sum FKL task, nuisance, and residual directions. The task direction equals $\Delta_t^{\mathrm{CTR}}$. \\
$r_t^{\mathrm{task}},r_t^{\mathrm{nuis}},r_t^{\FEC}$ & RKL task, nuisance, and Fisher-residual log-density contrasts. \\
$u_t^{\mathrm{task}},u_t^{\mathrm{nuis}},u_t^{\mathrm{FEC}}$ & Fisher-mapped task, nuisance, and residual logit directions. \\
$q_t^{\mathrm{pres},\RKL},\alpha_t^{\FKL},\alpha_t^{\RKL}$ & RKL evidence-presence midpoint and geometry-specific FEC coefficients. \\
$\epsilon_{\mathrm{proj}}$ & Ridge used in the Fisher projection denominator. \\
\midrule
\multicolumn{2}{@{}l@{}}{\textbf{Masks, objectives, and normalization}} \\

\midrule

$a_{i,t},m_{i,t}$ & Binary valid-response-token and admissible evidence masks, with $m_{i,t}\leq a_{i,t}$. \\
$N,Z_{\mathrm{evi}}$ & Response-token count and masked evidence-term normalizer. \\
$\Lgrpo,\Lref,\Levi,\Lverpo$ & Outcome, reference, evidence, and complete losses. \\
$\chi_{\mathrm{OPSD}},\chi_{\mathrm{adv}}$ & Binary switches for signed evidence correction and advantage modulation. \\
$\lambda_{\mathrm{ref}},\lambda_{\mathrm{evi}}$ & Coefficients for reference restoration and evidence correction. \\
$\FKL,\RKL,\FEC$ & Forward-KL geometry, reverse-KL geometry, and Fisher Evidence Contrast. \\
\midrule
\multicolumn{2}{@{}l@{}}{\textbf{ZPD controller and derivation quantities}} \\   

\midrule

$d_{i,t},\eta,J_{\mathrm{RL}}$ & Parameter-space evidence direction, local step size, and RL objective used by controller. \\
$\theta_{\mathrm{base}}$ & Hypothetical Actor state after the outcome and reference updates. \\
$B_{i,t},C_{i,t},w_{i,t}^{\mathrm{ideal}}$ & Directional benefit, curvature cost, and ideal token-wise acceptance. \\
$h_{i,t},c_{i,t},w_{i,t}$ & Alignment proxy, local-KL movement proxy, and stopped acceptance weight. \\
$\alpha_{\mathrm{cost}},\epsilon_{\mathrm{cost}}$ & Fisher-cost coefficient and positive denominator floor used by the ZPD controller. \\
\bottomrule
\end{tabularx}
\end{table}
\end{minipage}
\clearpage

\section{Algorithm}
\label{sec:algorithm}

\begin{algorithm}[H]
\caption{\method{} Actor Update under FKL Geometry (Detailed)}
\label{alg:verpo-zpd}
\small
\begin{algorithmic}[1]
\REQUIRE Current Actor $\pi_\theta$, EMA Teacher $\pi_{\bar\theta}$,
  dataset $\mathcal D$,
  variant $s\in\{\mathrm{Fix},\mathrm{CTR},\mathrm{FEC}\}$,
  switches $\chi_{\mathrm{OPSD}},\chi_{\mathrm{adv}}$
\ENSURE Updated Actor $\pi_\theta$

\REPEAT
  \STATE \textbf{Stage 1: Rollout Sampling and Outcome Signal}

  \STATE Sample a prompt batch from $\mathcal D$;
    freeze $\theta_{\mathrm{old}}\leftarrow\theta$.

  \STATE $\{y_i\}_{i\in\mathcal G(x)}
    \sim\pi_{\theta_{\mathrm{old}}}(\cdot\mid x)$
    for each sampled prompt $x$.

  \STATE Obtain verifier rewards $R_i$; set
    $\widehat A_i\leftarrow
      R_i-|\mathcal G(i)|^{-1}\sum_{j\in\mathcal G(i)}R_j$.
    \VERPOcomment{Outcome anchor}

  \STATE \textbf{Stage 2: Token-Level Evidence Correction}

  \STATE Score sampled prefixes with the current Actor
    under the original prompt.

  \STATE Replay the same prefixes under selected evidence contexts
    with the Teacher held fixed.

  \STATE $p_{i,t}\leftarrow
    \pi_\theta(\cdot\mid x_i,y_{i,<t})$.
    \VERPOcomment{Original-prompt distribution}

  \STATE Obtain stopped Teacher distributions
    $\{q_{i,t}^{\mathrm{ref}},q_{i,t}^0,q_{i,t}^+,q_{i,t}^-\}$
    via Eq.~\eqref{eq:app-teacher-replay}.
    \VERPOcomment{Same-prefix replay}

  \STATE \VERPOhl{verpoRed}{\Delta_{i,t}}
    $\leftarrow
    \begin{cases}
      q_{i,t}^+-q_{i,t}^0, & s=\mathrm{Fix},\\
      q_{i,t}^+-q_{i,t}^-, & s=\mathrm{CTR},\\
      \Delta_{i,t}^{\FEC}, & s=\mathrm{FEC},
    \end{cases}$
    with $\Delta_{i,t}^{\FEC}$ from Eq.~\eqref{eq:main-fec-direction}.
    \VERPOcomment[verpoRed]{Signed correction}

  \STATE \textbf{Stage 3: ZPD Weight Estimation}

  \STATE $\Delta_{i,t}\leftarrow\sg[\Delta_{i,t}]$,\quad
    $u_{i,t}\leftarrow\Delta_{i,t}$,\quad
    $h_{i,t}\leftarrow
      [\widehat A_i(e_{y_{i,t}}-p_{i,t})^\top u_{i,t}]_+$.
    \VERPOcomment[verpoBlue]{GRPO alignment}

  \STATE $c_{i,t}\leftarrow
    u_{i,t}^{\top}F(p_{i,t})u_{i,t}$.
    \VERPOcomment[verpoBlue]{Fisher movement cost}

  \STATE Form valid-token mask $a_{i,t}$ and
    evidence mask $m_{i,t}\leq a_{i,t}$.

  \STATE
    \VERPOhl{verpoOrange}{
      w^*_{i,t}\leftarrow\sg\!\left[
        \frac{h_{i,t}}{
          h_{i,t}+\alpha_{\mathrm{cost}}c_{i,t}
          +\epsilon_{\mathrm{cost}}}
      \right]
    },\quad
    $\widetilde w_{i,t}\leftarrow m_{i,t}w^*_{i,t}$.
    \VERPOcomment[verpoOrange]{Token-wise ZPD gate}

  \STATE \textbf{Stage 4: Two-Path Policy Update}

  \STATE Set
    \VERPOhl{verpoBlue}{
      \widehat A'_{i,t}\leftarrow
      \widehat A_i[1+\chi_{\mathrm{adv}}\lambda_{\mathrm{adv}}
      \sg(\widetilde w_{i,t})]
    }.
    \VERPOcomment[verpoBlue]{Advantage modulation}

  \STATE Set $N\leftarrow\sum_{i,t}a_{i,t}$ and
    $Z_{\mathrm{evi}}\leftarrow
      \max\!\left(1,\sum_{i,t}a_{i,t}m_{i,t}\right)$.
    \VERPOcomment{Separate normalizers}

  \STATE Compute reference and evidence losses via
    Eqs.~\eqref{eq:main-fkl-reference-loss}
    and~\eqref{eq:main-fkl-opsd-loss}.

  \STATE $\mathcal L_{\mathrm{total}}^{\FKL}\leftarrow
      \mathcal L_{\mathrm{GRPO}}(\widehat A')
      +\lambda_{\mathrm{ref}}\mathcal L_{\mathrm{ref}}^{\FKL}
      +{}$
    \VERPOhl{verpoRed}{
      \chi_{\mathrm{OPSD}}\lambda_{\mathrm{evi}}
      \mathcal L_{\mathrm{OPSD}}^{\FKL}(w^*)
    }.
    \VERPOcomment[verpoRed]{Evidence-loss path}

  \STATE $\theta\leftarrow\theta-\eta_\theta
    \nabla_\theta\mathcal L_{\mathrm{total}}^{\FKL}$.

  \STATE Update the EMA Teacher via Eq.~\eqref{eq:app-ema-teacher}.
    \VERPOcomment{After the Actor update}

\UNTIL{stopping criterion is reached}
\RETURN $\pi_\theta$
\end{algorithmic}
\end{algorithm}
\section{Preliminaries}
\label{sec:preliminaries}

To specify the quantities used by Algorithm~\ref{alg:verpo-zpd}, this appendix
defines rollout normalization, the outcome objective, frozen Teacher replay,
and local KL geometry. These conventions support the loss decompositions and
controller approximations in Appendix~\ref{sec:theoretical-analysis}.

\subsection{Autoregressive Policy}
\label{sec:app-notation}

To compare Actor and Teacher predictions at the same positions, we index
both distributions by the sampled response prefix. For each prompt, the frozen behavior Actor
$\pi_{\theta_{\mathrm{old}}}$ samples a group of responses. Index $i$ ranges
over responses, and $\mathcal G(i)$ denotes the responses generated for the
same prompt as response $i$. The current Actor factorizes as
\begin{equation}
\pi_\theta(y_i\mid x_i)
=\prod_{t=1}^{T_i}p_{i,t}(y_{i,t}),
\qquad
p_{i,t}=\pi_\theta(\cdot\mid x_i,y_{i,<t}).
\label{eq:app-autoregressive-policy}
\end{equation}
Each factor scores one observed token rather than a new Teacher completion.
When padded tensors are used, $a_{i,t}\in\{0,1\}$ is the valid-response-token
mask. Analytic sums over $t\leq T_i$ are equivalently restricted to
$a_{i,t}=1$. The binary evidence mask satisfies $m_{i,t}\leq a_{i,t}$, and the
normalizers are
\begin{equation}
N=\sum_{i,t}a_{i,t}=\sum_iT_i,
\qquad
Z_{\mathrm{evi}}
=\max\!\left(1,\sum_{i,t}a_{i,t}m_{i,t}\right).
\label{eq:app-normalizers}
\end{equation}
We assume a batch contains at least one valid response token, so $N>0$.
The reference and outcome losses average over these tokens. The evidence loss
averages over eligible tokens without renormalizing by acceptance weight.
If none are eligible, its numerator vanishes and the floor keeps its
normalizer defined.

To measure reward alignment, we use the unclipped local
GRPO~\citep{shao2024deepseekmath} score direction. For Actor logits $z_{i,t}$
and sampled-token one-hot vector $e_{y_{i,t}}$, this direction is
\begin{equation}
g_{i,t}^{\mathrm{RL}}
:=\widehat A_i\nabla_{z_{i,t}}\log p_{i,t}(y_{i,t})
=\widehat A_i(e_{y_{i,t}}-p_{i,t}).
\label{eq:app-grpo-score-direction}
\end{equation}
This vector converts the sequence advantage into a token-local comparison
direction for ZPD. It is not the gradient of the full clipped training loss,
which also depends on importance ratios and clipping.

\subsection{RLVR and Group-Relative Policy Optimization}
\label{sec:app-grpo-prelim}

To obtain the outcome signal used in Section~\ref{sec:overview}, we evaluate
each sampled response with a scalar verifier reward $R_i$.
Our GRPO~\citep{shao2024deepseekmath} variant centers this reward within its
prompt group without dividing by the within-group standard deviation:
\begin{equation}
\widehat A_i
=R_i-\frac{1}{|\mathcal G(i)|}
\sum_{j\in\mathcal G(i)}R_j.
\label{eq:app-grpo-advantage}
\end{equation}
The advantage measures performance relative to responses to the same prompt
and is broadcast across the response tokens. A group with identical rewards
has zero advantage for every response.

To account for the difference between the current and behavior policies,
define the sampled-token importance ratio as
\begin{equation*}
\rho_{i,t}(\theta)
=\frac{p_{i,t}(y_{i,t})}{p_{i,t}^{\mathrm{old}}(y_{i,t})}
\end{equation*}
The denominator remains fixed during the Actor update. With asymmetric clipping radii
$\epsilon_{\mathrm{low}},\epsilon_{\mathrm{high}}>0$, the outcome loss is
\begin{equation}
\mathcal L_{\mathrm{GRPO}}
=-\frac1N\sum_{i,t}a_{i,t}
\min\!\left\{
\rho_{i,t}(\theta)\widehat A_i,\,
\clip\!\left(
\rho_{i,t}(\theta),
1-\epsilon_{\mathrm{low}},
1+\epsilon_{\mathrm{high}}
\right)\widehat A_i
\right\}.
\label{eq:app-grpo-objective}
\end{equation}
The minimum limits the incentive to move a sampled-token ratio farther in
the advantage-favored direction after crossing its clipping threshold.
The valid-token mask excludes
padding, and $N$ fixes the averaging scale. Section~\ref{sec:two-paths} changes
the token advantage in this loss while retaining the same ratio and clipping rule.

\subsection{Feedback-Conditioned Teacher Replay}
\label{sec:app-teacher-replay}

To construct evidence comparisons without changing the sampled trajectory,
we replay its prefixes under different Teacher contexts.
SDPO~\citep{hubotter2026sdpo} and SRPO~\citep{li2026srpo} also use
feedback-conditioned replay for dense supervision. VERPO represents each
context condition $c$ through the stopped replay operator
\begin{equation}
q_{i,t}^{(c)}
=\sg\!\left[
\pi_{\bar\theta}
\!\left(
\cdot\mid\operatorname{reprompt}(x_i,c),y_{i,<t}
\right)
\right].
\label{eq:app-teacher-replay}
\end{equation}
The reprompt changes the Teacher's evidence context, while the Actor prefix
$y_{i,<t}$ stays fixed. Teacher forcing therefore scores the original
trajectory without generating a replacement completion. The stop-gradient
operator prevents this supervision branch from receiving Actor-loss gradients.

The Teacher parameters $\bar\theta$ are an exponential-moving-average copy of
the Actor across optimizer steps,
\begin{equation}
\bar\theta
\leftarrow
\gamma\bar\theta+(1-\gamma)\theta,
\qquad 0<\gamma<1,
\label{eq:app-ema-teacher}
\end{equation}
Here $\gamma$ retains the previous Teacher parameters and $1-\gamma$ weights
the newly updated Actor. The Teacher changes between optimizer steps but
remains frozen within each Actor update. Its parameters $\bar\theta$ are
distinct from the hypothetical Actor state $\theta_{\mathrm{base}}$ used
only in the controller derivation.

\subsection{Divergence and Local Information Geometry}
\label{sec:app-local-geometry}

To distinguish the two loss geometries, we fix the argument order of the
KL divergence~\citep{kullback1951information}. For positive, normalized
distributions over the vocabulary, our conventions are
\begin{equation}
\begin{aligned}
\KL(q\Vert p)
&=\sum_{v\in\mathcal V}q(v)\log\frac{q(v)}{p(v)},\\
D_{\FKL}(q,p)&=\KL(q\Vert p),
\qquad
D_{\RKL}(p,q)=\KL(p\Vert q).
\end{aligned}
\label{eq:app-kl-conventions}
\end{equation}
FKL averages the log-density ratio under the Teacher, whereas RKL averages
it under the Actor. We require positive support wherever these logarithms
are evaluated. Finite-logit softmax distributions satisfy this assumption
in exact arithmetic. The finite-support implementation is a separate
approximation described in Appendix~\ref{sec:app-method-hyperparameters}.

Let $z_{i,t}\in\mathbb R^{|\mathcal V|}$ be the Actor logits and
$p_{i,t}=\softmax(z_{i,t})$. For vocabulary items $u,v\in\mathcal V$, the
softmax score identity is
\begin{equation}
\frac{\partial\log p_{i,t}(u)}{\partial z_{i,t}(v)}
=\bm1[u=v]-p_{i,t}(v).
\label{eq:app-softmax-score}
\end{equation}
The Jacobian of the probabilities $p_{i,t}$ with respect to the logits,
not the Jacobian of their logarithms, gives the categorical Fisher matrix.
For an arbitrary vocabulary vector $b$,
\begin{equation}
F(p_{i,t})
=\diag(p_{i,t})-p_{i,t}p_{i,t}^\top,
\qquad
F(p_{i,t})\bm1=0,
\qquad
b^\top F(p_{i,t})b
=\operatorname{Var}_{V\sim p_{i,t}}[b(V)]\geq0.
\label{eq:app-fisher-prelim}
\end{equation}
Thus constant shifts of a log-density contrast do not change its induced logit
direction. A small logit displacement
$z_{i,t}\mapsto z_{i,t}+\epsilon b$ induces the local policy movement
\begin{equation}
\KL\!\left(
p_{i,t}\Vert\softmax(z_{i,t}+\epsilon b)
\right)
=\tfrac12\epsilon^2b^\top F(p_{i,t})b+O(\epsilon^3).
\label{eq:app-local-kl-prelim}
\end{equation}
For fixed finite logits and direction $b$, the quadratic term measures
movement of the Actor distribution to second order as $\epsilon$ tends to
zero. It supplies the controller's nonnegative movement proxy, not a
guarantee of reward improvement for a finite optimizer step.

\section{Theoretical Analysis}
\label{sec:theoretical-analysis}

This appendix provides the complete derivations behind the signed evidence
directions and token-wise controller in Section~\ref{sec:method}, using the
RLVR, Teacher-replay, and local-geometric conventions collected in
Appendix~\ref{sec:preliminaries}.
We first establish exact loss identities, then distinguish the local
approximations and chosen surrogate used to construct the controller.
Teacher distributions, evidence masks, acceptance weights, and constructed
FKL directions or RKL contrasts are stopped during Actor differentiation.
This includes the Actor-dependent FEC projection coefficients.

\subsection{Teacher Paths}
\label{sec:app-teacher-paths}

To vary evidence strength while preserving a distribution, we connect two
Teacher endpoints at the same rollout prefix $y_{<t}$. Only the evidence
condition changes. We assume normalized distributions on a finite vocabulary
and positive Actor probability on the Teacher support. For FKL, normalized endpoints imply
$\bm1^\top\Delta_t=0$ for $\Delta_t=q_{e,t}-q_{b,t}$. The arithmetic path is
\begin{equation}
q_{w,t}^{\FKL}
=(1-w)q_{b,t}+wq_{e,t}
=q_{b,t}+w\Delta_t,
\qquad 0\leq w\leq1.
\label{eq:app-fkl-path}
\end{equation}
The weight interpolates from the base endpoint to the enriched endpoint.
Convex mixing preserves normalization and nonnegativity for $0\leq w\leq1$,
while $\Delta_t$ records a signed probability displacement.

For RKL, assume positive support for both endpoints and define
\begin{equation}
r_t(v)=\log\frac{q_{e,t}(v)}{q_{b,t}(v)}.
\label{eq:app-rkl-contrast}
\end{equation}
This contrast measures relative density rather than an additive probability
change. To interpolate in these log-density coordinates, define
\begin{equation}
q_{w,t}^{\RKL}(v)
=\frac{q_{b,t}(v)^{1-w}q_{e,t}(v)^w}{Z_t(w)}
=\frac{q_{b,t}(v)e^{wr_t(v)}}{Z_t(w)},
\quad
Z_t(w)=\sum_{u\in\V}q_{b,t}(u)e^{wr_t(u)}.
\label{eq:app-rkl-path}
\end{equation}
The partition function $Z_t(w)$ normalizes the exponentially tilted base.
Both endpoints are recovered at $w=0$ and $w=1$. Unlike the arithmetic path,
the intermediate distributions mix log densities before normalization.

\subsection{Exact Forward-KL Decomposition}
\label{sec:app-fkl}

To recover the separation in Equation~\eqref{eq:main-fkl-decomposition},
we expand the FKL objective with the Teacher endpoints and $w$ held fixed.
We suppress the token index and assume positive Actor probability wherever
the Teacher assigns mass. The loss for this single prefix is
\begin{equation}
\mathcal L_{\FKL}(\theta;w)
=\KL(q_w^{\FKL}\Vert p_\theta).
\label{eq:app-fkl-objective}
\end{equation}
Only its cross-entropy term depends on the Actor. Substituting
$q_w^{\FKL}=q_b+w\Delta$ into that term gives
\begin{align*}
\mathcal L_{\FKL}(\theta;w)
&=\sum_{v\in\V}q_w^{\FKL}(v)\log q_w^{\FKL}(v)
-\sum_{v\in\V}q_w^{\FKL}(v)\log p_\theta(v)\\
&=\KL(q_b\Vert p_\theta)
-w\sum_{v\in\V}\Delta(v)\log p_\theta(v)
+C(q_w^{\FKL},q_b),
\end{align*}
where
\begin{equation*}
C(q_w^{\FKL},q_b)
=\sum_vq_w^{\FKL}(v)\log q_w^{\FKL}(v)
-\sum_vq_b(v)\log q_b(v).
\end{equation*}
Thus the exact decomposition is
\begin{equation}
\boxed{
\KL(q_w^{\FKL}\Vert p)
=\KL(q_b\Vert p)
-w\sum_{v\in\V}\Delta(v)\log p(v)
+C(q_w^{\FKL},q_b).}
\label{eq:app-fkl-decomposition}
\end{equation}
The first term restores the base distribution, and the second isolates the
signed evidence correction. The remainder $C$ contains only Teacher entropy
terms and has zero Actor gradient under the stated stop-gradient convention.
This is an exact identity, without a small-step approximation.

To verify the correction sign, differentiate with respect to the logits of
$p=\softmax(z)$ using
\begin{equation*}
\frac{\partial\log p(u)}{\partial z(v)}
=\bm1[u=v]-p(v).
\end{equation*}
Therefore
\begin{align*}
\frac{\partial\mathcal L_{\FKL}}{\partial z(v)}
&=-\sum_{u\in\V}q_w^{\FKL}(u)
\bigl(\bm1[u=v]-p(v)\bigr)\\
&=p(v)-q_w^{\FKL}(v),
\end{align*}
and the exact logit descent direction decomposes as
\begin{equation}
\boxed{
-\nabla_z\mathcal L_{\FKL}
=\underbrace{q_b-p}_{\text{base restoration}}
+\underbrace{w\Delta}_{\text{evidence correction}}.}
\label{eq:app-fkl-gradient}
\end{equation}
The base contribution points from the Actor toward $q_b$, while $w\Delta$
adds the accepted Teacher displacement. Retaining only this second channel
gives the signed evidence loss derived from OPSD~\citep{zhao2026opsd}:
\begin{equation}
\mathcal L_{\mathrm{evi}}^{\FKL}(w)
=-w\sum_{v\in\V}\Delta(v)\log p(v),
\qquad
-\nabla_z\mathcal L_{\mathrm{evi}}^{\FKL}=w\Delta.
\label{eq:app-fkl-correction}
\end{equation}
The zero-sum property of $\Delta$ cancels the softmax normalization term in
the gradient. A separate reference loss can therefore use
$q^{\mathrm{ref}}$ even when the evidence comparison starts at a different
base $q_b$.

\subsection{Reverse-KL Counterpart}
\label{sec:app-rkl}

To obtain the counterpart of the FKL separation, we substitute the geometric
Teacher path into RKL. Assume positive Actor and Teacher distributions on
the vocabulary, with $r$ and $w$ stopped. The logarithm of the normalized
path separates into a base density, a weighted contrast, and a scalar
normalizer:
\begin{align*}
\mathcal L_{\RKL}(\theta;w)
&=\KL(p_\theta\Vert q_w^{\RKL})\\
&=\sum_{v\in\V}p(v)
\left[\log p(v)-\log q_b(v)-wr(v)+\log Z(w)\right]\\
&=\KL(p\Vert q_b)-w\mathbb E_{v\sim p}[r(v)]+\log Z(w).
\end{align*}
Hence
\begin{equation}
\boxed{
\KL(p\Vert q_w^{\RKL})
=\KL(p\Vert q_b)
-w\mathbb E_{v\sim p}[r(v)]
+\log Z(w).}
\label{eq:app-rkl-decomposition}
\end{equation}
The base term retains the original Teacher anchor. The contrast term changes
the Actor's expected log-density preference, and $\log Z(w)$ completes the
loss identity without contributing an Actor gradient.

To identify the resulting logit direction, apply the probability Jacobian
$F(p)$ to the derivative of RKL. For a fixed positive Teacher $q$,
\begin{align*}
\nabla_z\KL(p\Vert q)
&=F(p)\bigl(\log p-\log q+\bm1\bigr)\\
&=F(p)(\log p-\log q),
\end{align*}
because $F(p)\bm1=0$. Substituting
$\log q_w^{\RKL}=\log q_b+wr-\log Z(w)$ yields
\begin{equation}
\boxed{
-\nabla_z\mathcal L_{\RKL}
=\underbrace{F(p)(\log q_b-\log p)}_{\text{base restoration}}
+\underbrace{w\xi}_{\text{evidence correction}},
\qquad
\xi=F(p)r.}
\label{eq:app-rkl-gradient}
\end{equation}
Here $\xi$ denotes the RKL evidence direction, written as $u$ in the unified
controller. Unlike the FKL displacement, it depends on the Actor through
$F(p)$. Writing $\bar r=\mathbb E_{v\sim p}[r(v)]$ exposes this dependence:
\begin{equation}
\xi(v)=p(v)\bigl(r(v)-\bar r\bigr),
\qquad
\bm1^\top\xi=0.
\label{eq:app-rkl-elementwise}
\end{equation}
The Fisher map centers the log contrast under the Actor and weights each
coordinate by its current probability. Thus $\xi$ sums to zero but is not
the probability displacement $q_e-q_b$. Its signed loss is
\begin{equation}
\mathcal L_{\mathrm{evi}}^{\RKL}(w)
=\log Z(w)-w\mathbb E_{v\sim p}[r(v)],
\qquad
-\nabla_z\mathcal L_{\mathrm{evi}}^{\RKL}=wF(p)r.
\label{eq:app-rkl-correction}
\end{equation}
Differentiation acts through the expectation under $p$, not through the
stopped contrast. The resulting $F(p)r$ supplies the direction used by the
RKL controller and the complete gradient in
Appendix~\ref{sec:app-complete-objective}.

\subsection{Fixed, Contrastive, and Fisher Evidence Contrast}
\label{sec:app-directions}

To instantiate the evidence channel in Section~\ref{sec:fec}, we specify
the endpoints and the FEC residual in both geometries. All branches replay
the same prefix. The RKL constructions require positive Teacher support,
and every projection is computed at the current Actor distribution before
its coefficient is stopped.

\paragraph{Fixed Evidence.}
The Fixed construction uses the evidence-conditioned replay of
SDPO~\citep{hubotter2026sdpo} to define the endpoints $q_b=q^0$ and $q_e=q^+$:
\begin{equation}
\Delta_t^{\mathrm{Fix}}=q_t^+-q_t^0,
\qquad
r_t^{\mathrm{Fix}}=\log q_t^+-\log q_t^0,
\qquad
u_t^{\mathrm{Fix},\RKL}=F(p_t)r_t^{\mathrm{Fix}}.
\label{eq:app-fixed-direction}
\end{equation}
This comparison includes both the change
associated with correct evidence and the change associated with presenting
evidence at all. The observed reward hacking for this tested direction is
reported with Table~\ref{tab:component-ablation}, not implied by this identity.

\paragraph{Contrastive Evidence.}
The endpoint comparison follows RLCSD~\citep{pan2026rlcsd}.
For $K\geq1$ incorrect-evidence Teachers, define
$q_t^-=K^{-1}\sum_{k=1}^Kq_t^{-,k}$. The construction uses
$q_b=q^-$ and $q_e=q^+$, so
\begin{equation}
\Delta_t^{\mathrm{CTR}}=q_t^+-q_t^-,
\qquad
r_t^{\mathrm{CTR}}=\log q_t^+-\log q_t^-,
\qquad
u_t^{\mathrm{CTR},\RKL}=F(p_t)r_t^{\mathrm{CTR}}.
\label{eq:app-contrastive-direction}
\end{equation}
The incorrect branches are averaged in probability space before taking a
logarithm for RKL. Subtraction cancels additive shifts shared by the two
endpoints, but it retains context effects that differ between them.

\paragraph{Forward-KL FEC.}
For normalized Teacher branches and a fixed Actor Fisher matrix, define the
task contrast and evidence-presence proxy. The target claim concerns the
residual Fisher inner product, without assuming that the proxy contains only
task-irrelevant information. In probability coordinates, the directions are
\begin{equation*}
\Delta_t^{\mathrm{task}}=q_t^+-q_t^- ,
\qquad
\Delta_t^{\mathrm{nuis}}=\tfrac12(q_t^++q_t^-)-q_t^0.
\end{equation*}
Both vectors sum to zero. The task vector equals $\Delta_t^{\mathrm{CTR}}$
in the main text, while the midpoint comparison measures evidence presence.
For the Fisher inner product
$\langle a,b\rangle_{F(p_t)}=a^\top F(p_t)b$, the projection coefficient and
residual are
\begin{equation}
\boxed{
\alpha_t^{\FKL}
=\frac{\langle\Delta_t^{\mathrm{task}},
\Delta_t^{\mathrm{nuis}}\rangle_{F(p_t)}}
{\langle\Delta_t^{\mathrm{nuis}},
\Delta_t^{\mathrm{nuis}}\rangle_{F(p_t)}+\epsilon_{\mathrm{proj}}},
\qquad
\Delta_t^{\FEC}
=\Delta_t^{\mathrm{task}}-\alpha_t^{\FKL}\Delta_t^{\mathrm{nuis}}.}
\label{eq:app-fec-fkl}
\end{equation}
The coefficient $\alpha_t^{\FKL}$ is the FKL coefficient denoted
$\alpha^{\FEC}$ in Section~\ref{sec:fec}. It subtracts a component along the
measured nuisance direction. To determine what regularization leaves behind,
substitute this coefficient into the residual inner product:
\begin{equation}
\left\langle\Delta_t^{\FEC},\Delta_t^{\mathrm{nuis}}\right\rangle_{F(p_t)}
=\frac{\epsilon_{\mathrm{proj}}}
{\langle\Delta_t^{\mathrm{nuis}},\Delta_t^{\mathrm{nuis}}\rangle_{F(p_t)}
+\epsilon_{\mathrm{proj}}}
\left\langle\Delta_t^{\mathrm{task}},\Delta_t^{\mathrm{nuis}}\right\rangle_{F(p_t)}.
\label{eq:app-fec-fkl-residual}
\end{equation}
For $\epsilon_{\mathrm{proj}}>0$, the remaining inner product equals the
original one times a factor in $(0,1]$. Exact orthogonality follows in the
unregularized case only when the denominator is nonzero.
The residual still sums to zero, but adding it to a base distribution need
not preserve nonnegativity. We therefore use the stopped residual directly
in the signed loss:
\begin{equation}
\mathcal L_{\FEC,t}^{\FKL}
=-w_t\sum_{v\in\V}\Delta_t^{\FEC}(v)\log p_t(v),
\qquad
-\nabla_{z_t}\mathcal L_{\FEC,t}^{\FKL}=w_t\Delta_t^{\FEC}.
\label{eq:app-fec-fkl-loss}
\end{equation}
The stated gradient requires stopping the entire residual, including its
Actor-dependent coefficient. It recovers the direction in
Equation~\eqref{eq:main-fec-direction} without claiming a valid arithmetic
Teacher endpoint for FEC.

\paragraph{Reverse-KL FEC.}
To apply the same projection idea in RKL geometry, first measure evidence
presence using positive Teacher branches and the normalized geometric midpoint
\begin{equation*}
q_t^{\mathrm{pres},\RKL}(v)
=\frac{\sqrt{q_t^+(v)q_t^-(v)}}
{\sum_{u\in\V}\sqrt{q_t^+(u)q_t^-(u)}}
\end{equation*}
to measure evidence presence in log-density coordinates. Define
\begin{align*}
r_t^{\mathrm{task}}&=\log q_t^+-\log q_t^-,
&u_t^{\mathrm{task}}&=F(p_t)r_t^{\mathrm{task}},\\
r_t^{\mathrm{nuis}}&=\log q_t^{\mathrm{pres},\RKL}-\log q_t^0,
&u_t^{\mathrm{nuis}}&=F(p_t)r_t^{\mathrm{nuis}}.
\end{align*}
These log contrasts are mapped into logit directions before projection.
The midpoint normalization does not change $u_t^{\mathrm{nuis}}$ because
$F(p_t)\bm1=0$. Projecting in the same Fisher inner product gives
\begin{equation}
\boxed{
\alpha_t^{\RKL}
=\frac{(u_t^{\mathrm{task}})^\top F(p_t)u_t^{\mathrm{nuis}}}
{(u_t^{\mathrm{nuis}})^\top F(p_t)u_t^{\mathrm{nuis}}
+\epsilon_{\mathrm{proj}}},
\quad
r_t^{\FEC}=r_t^{\mathrm{task}}-\alpha_t^{\RKL}r_t^{\mathrm{nuis}},
\quad
u_t^{\FEC}=F(p_t)r_t^{\FEC}.}
\label{eq:app-fec-rkl}
\end{equation}
Linearity of $F(p_t)$ gives
$u_t^{\FEC}=u_t^{\mathrm{task}}-\alpha_t^{\RKL}u_t^{\mathrm{nuis}}$.
Its residual Fisher inner product has the same ridge-dependent form as
Equation~\eqref{eq:app-fec-fkl-residual}, with $u$ replacing $\Delta$.
We stop the constructed contrast $r_t^{\FEC}$, including $\alpha_t^{\RKL}$,
before Actor differentiation. Using $q_t^-$ as the base then defines
\begin{equation}
q_{w,t}^{\RKL,\FEC}(v)
=\frac{q_t^-(v)e^{w_tr_t^{\FEC}(v)}}{Z_t^{\FEC}(w_t)},
\qquad
Z_t^{\FEC}(w)=\sum_{u\in\V}q_t^-(u)e^{wr_t^{\FEC}(u)}.
\label{eq:app-fec-rkl-path}
\end{equation}
Exponential tilting preserves a positive normalized distribution for finite
contrasts. Its endpoint at $w=1$ need not equal $q_t^+$ after projection.
The corresponding signed correction is
\begin{equation}
\mathcal L_{\FEC,t}^{\RKL}
=\log Z_t^{\FEC}(w_t)
-w_t\mathbb E_{v\sim p_t}[r_t^{\FEC}(v)],
\qquad
-\nabla_{z_t}\mathcal L_{\FEC,t}^{\RKL}=w_tu_t^{\FEC}.
\label{eq:app-fec-rkl-loss}
\end{equation}
The Fisher map in this gradient comes from differentiating the Actor
expectation. No gradient flows through the projection construction.
For controller evaluation, the resulting direction and acceptance weight
are treated as fixed inputs to the Actor update.

\subsection{Ideal Token-Wise Controller}
\label{sec:app-zpd-ideal}

To connect acceptance to local reward improvement, we analyze one fixed
evidence direction at a time. Let $G_{i,t}=\partial z_{i,t}/\partial\theta$
and $d_{i,t}=G_{i,t}^\top u_{i,t}$ be its parameter-space descent direction.
The hypothetical Actor state
$\theta_{\mathrm{base}}$ absorbs the outcome and reference updates for this
local analysis. It differs from the EMA Teacher state $\bar\theta$ and does
not require separate optimizer steps. At a fixed local step size $\eta>0$,
applying the correction with strength $w$ produces
\begin{equation}
\theta^+(w)=\theta_{\mathrm{base}}+\eta wd_{i,t}.
\label{eq:app-zpd-hypothetical-params}
\end{equation}
This defines a one-dimensional family of candidate Actor states rather than
an additional training loop. The ideal weight selects the state with the
largest RL objective:
\begin{equation}
\boxed{
w_{i,t}^{\mathrm{ideal}}
=\arg\max_{0\leq w\leq1}
J_{\mathrm{RL}}(\theta_{\mathrm{base}}+\eta wd_{i,t}).}
\label{eq:app-zpd-ideal-problem}
\end{equation}
Evaluating this objective for each token is not the deployed controller.
To expose its local benefit and cost, assume that $J_{\mathrm{RL}}$ has
locally bounded third derivatives along the fixed direction. Taylor
expansion for a sufficiently small step gives
\begin{align*}
J_{\mathrm{RL}}(\theta_{\mathrm{base}}+\eta wd_{i,t})
-J_{\mathrm{RL}}(\theta_{\mathrm{base}})
&=\eta w\nabla J_{\mathrm{RL}}(\theta_{\mathrm{base}})^\top d_{i,t}\\
&\quad+\tfrac12\eta^2w^2d_{i,t}^\top
\nabla^2J_{\mathrm{RL}}(\theta_{\mathrm{base}})d_{i,t}
+O((\eta w)^3).
\end{align*}
Define the directional benefit and curvature cost
\begin{equation}
B_{i,t}=\nabla J_{\mathrm{RL}}(\theta_{\mathrm{base}})^\top d_{i,t},
\qquad
C_{i,t}=-d_{i,t}^\top\nabla^2J_{\mathrm{RL}}(\theta_{\mathrm{base}})d_{i,t}.
\label{eq:app-zpd-exact-benefit-cost}
\end{equation}
The sign of $B_{i,t}$ determines first-order improvement, while positive
$C_{i,t}$ denotes negative reward curvature along the direction.
Under this additional condition, the quadratic model is
$\eta wB_{i,t}-\tfrac12\eta^2w^2C_{i,t}$. Its constrained optimum is
\begin{equation}
\boxed{
w_{i,t}^{\mathrm{quad}}
=\clip\!\left(\frac{[B_{i,t}]_+}{\eta C_{i,t}},0,1\right).}
\label{eq:app-zpd-quadratic-solution}
\end{equation}
This maximizes the truncated quadratic, not the original RL objective.
When $C_{i,t}\leq0$, the diminishing-return argument does not apply.
The practical controller therefore replaces both quantities with computable
proxies and uses a separately chosen bounded surrogate.

\subsection{Benefit and Local-KL Cost Proxies}
\label{sec:app-zpd-proxies}

To avoid computing the true reward gradient and Hessian, we construct the
proxies in Equation~\eqref{eq:zpd-proxies-main}. At sampled token $y_{i,t}$,
the unclipped local GRPO~\citep{shao2024deepseekmath} score gives the estimator
\begin{equation*}
\nabla_\theta J_{i,t}
\approx G_{i,t}^\top\widehat A_i(e_{y_{i,t}}-p_{i,t}).
\end{equation*}
This is the parameter-space pullback of the local logit direction in
Equation~\eqref{eq:app-grpo-score-direction}. It supplies a comparison
direction without replacing the clipped training objective.
Substituting $d_{i,t}=G_{i,t}^\top u_{i,t}$ into the exact directional benefit
gives
\begin{align*}
B_{i,t}
&\approx\widehat A_i(e_{y_{i,t}}-p_{i,t})^\top
G_{i,t}G_{i,t}^\top u_{i,t}\\
&\approx\kappa_{i,t}\widehat A_i
(e_{y_{i,t}}-p_{i,t})^\top u_{i,t},
\qquad \kappa_{i,t}>0,
\end{align*}
where the second line uses the local isotropic approximation
$G_{i,t}G_{i,t}^\top\approx\kappa_{i,t}I$. Absorbing the unknown positive
scale and rejecting negative alignment yields
\begin{equation}
\boxed{
h_{i,t}
=\left[\widehat A_i(e_{y_{i,t}}-p_{i,t})^\top u_{i,t}\right]_+.}
\label{eq:app-zpd-benefit-proxy}
\end{equation}
The isotropic approximation discards anisotropy in the logit Jacobian and
the score estimator discards interactions with other tokens. Consequently,
$h_{i,t}$ measures local alignment rather than verified reward improvement.
Its positive part rejects nonpositive alignment under this proxy.

The reward Hessian is generally indefinite, so $C_{i,t}>0$ need not hold for
the exact quadratic. Replace the negative Hessian along this update by a
positive-semidefinite Fisher surrogate, using the local information geometry
underlying natural-gradient methods~\citep{amari1998natural}:
\begin{equation}
-\nabla_\theta^2J_{\mathrm{RL}}(\theta_{\mathrm{base}})
\approx\beta_{\mathrm{curv}} F_\theta,
\qquad
F_\theta=G_{i,t}^\top F(p_{i,t})G_{i,t},
\qquad \beta_{\mathrm{curv}}>0.
\label{eq:app-zpd-curvature-replacement}
\end{equation}
The positive scale $\beta_{\mathrm{curv}}$ belongs only to this curvature
approximation and differs from the EMA decay $\gamma$. This replacement is
a modeling choice, not an identity for the reward Hessian.
Reusing the isotropic approximation gives
\begin{align*}
C_{i,t}
&\approx\beta_{\mathrm{curv}} d_{i,t}^\top F_\theta d_{i,t}\\
&=\beta_{\mathrm{curv}}(G_{i,t}d_{i,t})^\top F(p_{i,t})(G_{i,t}d_{i,t})\\
&\approx\beta_{\mathrm{curv}}\kappa_{i,t}^2u_{i,t}^\top F(p_{i,t})u_{i,t}.
\end{align*}
This quadratic form is the leading policy movement by
Equation~\eqref{eq:app-local-kl-prelim} with $b=u_{i,t}$.
Define the Fisher movement term and absorb the unknown positive curvature scale
into the active cost coefficient:
\begin{equation}
\boxed{
c_{i,t}=u_{i,t}^\top F(p_{i,t})u_{i,t},
\qquad
u_{i,t}^\top F(p_{i,t})u_{i,t}
=\operatorname{Var}_{V\sim p_{i,t}}[u_{i,t}(V)]\geq0.}
\label{eq:app-zpd-cost-proxy}
\end{equation}
The variance identity establishes nonnegativity without assumptions on the
reward Hessian. For a small logit step $\delta u_{i,t}$, the leading KL
movement is $\tfrac12\delta^2c_{i,t}$. The coefficient
$\alpha_{\mathrm{cost}}$ controls this proxy's influence in the deployed
weight, without estimating the omitted curvature scale.

\subsection{Surrogate Acceptance Optimization and Properties}
\label{sec:app-zpd-surrogate}

To turn the two proxies into acceptance, we use the surrogate introduced in
Section~\ref{sec:zpd}. Suppress token indices and assume
finite $h\geq0$, $c\geq0$, $\alpha_{\mathrm{cost}}\geq0$, and
$\epsilon_{\mathrm{cost}}>0$. With
$\Lambda=\alpha_{\mathrm{cost}}c+\epsilon_{\mathrm{cost}}$, define
\begin{equation}
f(w)=h\log w+\Lambda\log(1-w),
\qquad w\in(0,1).
\label{eq:app-zpd-surrogate-objective}
\end{equation}
For positive benefit, the first term penalizes zero acceptance and the
second penalizes full acceptance. This choice retains a benefit-versus-cost
tradeoff but is not derived as an exact reward objective. For $h>0$,
$f''(w)=-h/w^2-\Lambda/(1-w)^2<0$, so $f$ is strictly concave. The first-order
condition is
\begin{equation*}
\frac{h}{w}-\frac{\Lambda}{1-w}=0,
\end{equation*}
whose unique global maximizer is
\begin{equation}
\boxed{
w^\star
=\frac{h}{h+\Lambda}
=\frac{h}
{h+\alpha_{\mathrm{cost}}c+\epsilon_{\mathrm{cost}}}
\in[0,1).}
\label{eq:app-zpd-rational-controller}
\end{equation}
For $h=0$, the function reduces to $\Lambda\log(1-w)$ and has no maximizer
in the open interval. Its supremum occurs as $w$ tends to zero, so we extend
the solution continuously to $w^\star=0$. The deployed controller sets
$w_{i,t}=\sg[w^\star_{i,t}]$, using $w_{i,t}^{*}$ for the same stopped
weight in the main text. These bounds require the default positive-part
benefit. They do not apply to an ablation that inserts an unclamped signed
benefit into the same fraction.

For $h>0$, the acceptance odds provide another interpretation of this solution:
\begin{equation}
\frac{w}{1-w}
=\frac{h}{\alpha_{\mathrm{cost}}c+\epsilon_{\mathrm{cost}}},
\qquad
\operatorname{logit}(w)
=\log h-\log(\alpha_{\mathrm{cost}}c+\epsilon_{\mathrm{cost}}).
\label{eq:app-zpd-logodds}
\end{equation}
The log-odds expression requires positive benefit, whereas the rational rule
also covers the zero-benefit boundary. To verify monotonicity before stopping
the weight, differentiate the rational expression:
\begin{equation*}
\frac{\partial w^\star}{\partial h}
=\frac{\Lambda}{(h+\Lambda)^2}>0,\qquad
\frac{\partial w^\star}{\partial c}
=-\frac{\alpha_{\mathrm{cost}}h}{(h+\Lambda)^2}\leq0,\qquad
\frac{\partial w^\star}{\partial\alpha_{\mathrm{cost}}}
=-\frac{hc}{(h+\Lambda)^2}\leq0.
\end{equation*}
Cost strictly reduces acceptance when $h>0$ and
$\alpha_{\mathrm{cost}}>0$. At $h=0$, the controller rejects the direction
regardless of cost. The positive floor keeps acceptance below one for
finite inputs.
PPO~\citep{schulman2017ppo} clips a policy probability ratio, whereas ZPD
computes a stopped scale from evidence alignment and movement cost.
Neither these monotonicity properties nor the local proxies guarantee a
reward increase. Reference restoration remains ungated in both update paths.

\subsection{Multiplicative Advantage Modulation}
\label{sec:app-advantage-modulation}

To verify Path~1 in Section~\ref{sec:two-paths}, we insert acceptance into
the outcome advantage instead of adding an evidence gradient.
With the path enabled, define
\begin{equation}
\widetilde w_{i,t}=m_{i,t}w_{i,t},
\qquad
s_{i,t}=1+\lambda_{\mathrm{adv}}\,\sg\!\left(\widetilde w_{i,t}\right),
\qquad \lambda_{\mathrm{adv}}\geq0,
\label{eq:app-advantage-modulation-scale}
\end{equation}
and define the token-level advantage override
\begin{equation}
\boxed{
\widehat A'_{i,t}=s_{i,t}\widehat A_i
=\widehat A_i\left[
1+\lambda_{\mathrm{adv}}\,\sg\!\left(\widetilde w_{i,t}\right)
\right].}
\label{eq:app-advantage-modulation}
\end{equation}
Here $\widehat A_i$ is the original group-relative sequence advantage,
repeated over valid response tokens, while $s_{i,t}$ is token dependent. The
stop-gradient prevents the current loss from differentiating through the
controller to change this scale. The scale can still change when it is
recomputed at a later Actor state.

To check its interaction with clipping, write the token-level loss inherited
from PPO~\citep{schulman2017ppo} using the same importance ratio and asymmetric
radii as Equation~\eqref{eq:app-grpo-objective}:
\begin{equation}
\ell^{\mathrm{PPO}}_{i,t}(A)
=-\min\!\left(
\rho_{i,t}(\theta)A,
\clip(\rho_{i,t}(\theta),1-\epsilon_{\mathrm{low}},1+\epsilon_{\mathrm{high}})A
\right).
\label{eq:app-advantage-modulation-ppo}
\end{equation}
For the default nonnegative controller, $s_{i,t}>0$ is fixed during the
update, so positive scaling factors out of both branches:
\begin{equation}
\ell^{\mathrm{PPO}}_{i,t}(\widehat A'_{i,t})
=s_{i,t}\ell^{\mathrm{PPO}}_{i,t}(\widehat A_i),
\qquad
\nabla_\theta\ell^{\mathrm{PPO}}_{i,t}(\widehat A'_{i,t})
=s_{i,t}\nabla_\theta\ell^{\mathrm{PPO}}_{i,t}(\widehat A_i).
\label{eq:app-advantage-modulation-gradient}
\end{equation}
Thus positive scaling preserves the advantage sign and the selected clipping
branch, while scaling its gradient wherever the loss is differentiable.
At a clipping boundary, the same relation holds for a consistently chosen
subgradient. Ineligible tokens and zero acceptance both give $s_{i,t}=1$.
The mask therefore limits modulation without changing the reference loss.

\subsection{Complete Objective and Exact Logit Gradients}
\label{sec:app-complete-objective}

To assemble the training update, we combine the two channels with the
normalizers in Equation~\eqref{eq:app-normalizers}. Assume $N>0$, binary
$m_{i,t}\leq a_{i,t}$, and stopped Teacher distributions, constructed
directions, masks, and acceptance weights.
Here $\Levi$ denotes the signed evidence loss written as
$\mathcal L_{\mathrm{OPSD}}$ in the main text. The FKL terms are
\begin{align}
\Lref^{\FKL}
&=\frac1N\sum_{i,t}a_{i,t}\KL(q_{i,t}^{\mathrm{ref}}\Vert p_{i,t}),
\nonumber\\
\Levi^{\FKL}
&=-\frac1{Z_{\mathrm{evi}}}\sum_{i,t}m_{i,t}w_{i,t}
\sum_{v\in\V}\Delta_{i,t}(v)\log p_{i,t}(v).
\label{eq:app-complete-fkl-losses}
\end{align}
The reference channel averages over valid response tokens, while the evidence
channel averages over the eligible subset. Because $m_{i,t}\leq a_{i,t}$,
an additional factor $a_{i,t}$ in the evidence numerator would be redundant.
For RKL, assume positive support and use the selected stopped log contrast
$r_{i,t}$, including the residual contrast when FEC is enabled:
\begin{align}
\Lref^{\RKL}
&=\frac1N\sum_{i,t}a_{i,t}\KL(p_{i,t}\Vert q_{i,t}^{\mathrm{ref}}),
\nonumber\\
\Levi^{\RKL}
&=\frac1{Z_{\mathrm{evi}}}\sum_{i,t}m_{i,t}
\left[\log Z_{i,t}(w_{i,t})
-w_{i,t}\mathbb E_{v\sim p_{i,t}}[r_{i,t}(v)]\right].
\label{eq:app-complete-rkl-losses}
\end{align}
The RKL partition term preserves the loss identity but contributes no Actor
gradient. The reference Teacher remains evidence-free even when the selected
contrast uses $q^-$ as its base.
For binary path switches $\chi_{\mathrm{OPSD}},\chi_{\mathrm{adv}}$, define
$\widetilde w_{i,t}=m_{i,t}w_{i,t}$ and
$\widehat A'_{i,t}=\widehat A_i[1+\chi_{\mathrm{adv}}\lambda_{\mathrm{adv}}
\sg(\widetilde w_{i,t})]$. Combining these terms with
GRPO~\citep{shao2024deepseekmath} gives Equation~\eqref{eq:verpo-objective}
under FKL and its geometry-matched RKL counterpart. Enabling only
$\chi_{\mathrm{adv}}$ gives Path~1, enabling only $\chi_{\mathrm{OPSD}}$
gives Path~2, and enabling both gives the joint setting.
Write $\Lverpo$ for the complete loss denoted $\mathcal L_{\mathrm{total}}$
in the main text. Differentiating the three terms separately gives the
following token-logit descent directions, with the outcome loss evaluated
using the stopped modulated advantage. For compactness, write
$\lambda_{\mathrm{evi}}^{\star}=\chi_{\mathrm{OPSD}}\lambda_{\mathrm{evi}}$:
\begin{align}
-\nabla_{z_{i,t}}\Lverpo^{\FKL}
&=-\nabla_{z_{i,t}}\Lgrpo(\widehat A')
+\frac{\lambda_{\mathrm{ref}}a_{i,t}}{N}
(q_{i,t}^{\mathrm{ref}}-p_{i,t})
+\frac{\lambda_{\mathrm{evi}}^{\star}}{Z_{\mathrm{evi}}}
m_{i,t}w_{i,t}\Delta_{i,t},
\nonumber\\
-\nabla_{z_{i,t}}\Lverpo^{\RKL}
&=-\nabla_{z_{i,t}}\Lgrpo(\widehat A')
+\frac{\lambda_{\mathrm{ref}}a_{i,t}}{N}F(p_{i,t})
(\log q_{i,t}^{\mathrm{ref}}-\log p_{i,t})
+\frac{\lambda_{\mathrm{evi}}^{\star}}{Z_{\mathrm{evi}}}
m_{i,t}w_{i,t}F(p_{i,t})r_{i,t}.
\label{eq:app-complete-logit-gradients}
\end{align}
These exact logit gradients recover the main-text FKL update and its RKL
counterpart under the stated stop-gradient and support assumptions.
The reference term contains neither acceptance weights nor path switches.
ZPD scales the outcome gradient through the advantage in Path~1 and the
signed evidence direction directly in Path~2. The two routes therefore share
acceptance without sharing a gradient channel.

\section{Additional Results}
\label{sec:app-additional-results}

To examine the scope of the main results, this appendix reports joint-path
performance, trajectory routing, offline token weights, and evidence-loss
sensitivity, followed by a support-mass audit of vocabulary truncation.
The comparisons extend Table~\ref{tab:main-results} and the
controller analysis in Section~\ref{further:token_level_fec_weight}.

\subsection{Joint Loss Weighting and Advantage Modulation}
\label{sec:app-joint-paths}

\paragraph{Joint Activation.}
To test additive gains, we enabled both switches in
Section~\ref{sec:two-paths} as VERPO-LW+AM. Under the main evaluation
protocol, Table~\ref{tab:joint-path-results} shows that joint activation
falls below the stronger single-path setting across the evaluated backbones.
These endpoint results show no additive average gains but do not explain
the decrease.

\begin{table}[h]
\centering
\small
\caption{Acc/mean@16 of VERPO-LW+AM
across three model backbones. The Llama row uses all-trajectory correction.
All entries are percentages, with percentage symbols omitted.
Average is the unweighted five-task mean. Higher is better.}
\label{tab:joint-path-results}
\renewcommand{\arraystretch}{1.04}
\setlength{\tabcolsep}{4.2pt}
\begin{tabular*}{\textwidth}{@{\extracolsep{\fill}}lrrrrrr@{}}
\toprule
Model & Biology & Chemistry & Materials & Physics & Tool Use & Average \\
\midrule
Qwen3-4B & 60.16 & 70.17 & 77.26 & 68.98 & 62.56 & 67.83 \\
Qwen3-8B & 61.78 & 75.62 & 74.62 & 68.82 & 67.15 & 69.60 \\
Llama-3.2-1B & 55.38 & 70.41 & 56.38 & 44.29 & 49.26 & 55.14 \\
\bottomrule
\end{tabular*}
\end{table}

\begin{figure}[htbp]
\centering
\newsavebox{\pathannealingtablebox}
\sbox{\pathannealingtablebox}{%
\begin{minipage}{0.34\textwidth}
\centering
\small
\renewcommand{\arraystretch}{1.35}
\setlength{\tabcolsep}{0pt}
% Generated from fig/data/path_annealing.json.
\begin{tabular*}{\linewidth}{@{\extracolsep{\fill}}lrrr@{}}
\toprule
Task & 50 & 150 & 200 \\
\midrule
Biology & 37.50 & 15.12 & 19.12 \\
Chemistry & 47.29 & 62.41 & 62.62 \\
Materials & 35.17 & 19.02 & 27.26 \\
Physics & 40.08 & 38.05 & 36.48 \\
Tool Use & 44.49 & 45.68 & 48.25 \\
\bottomrule
\end{tabular*}

\end{minipage}}
\newlength{\pathannealingpanelheight}
\setlength{\pathannealingpanelheight}{%
  \dimexpr\ht\pathannealingtablebox+\dp\pathannealingtablebox\relax}
\typeout{Path annealing panel height: \the\pathannealingpanelheight}
\captionsetup[subfigure]{position=top,font={small,normalfont},labelfont=bf,
  justification=centering,singlelinecheck=false,skip=5pt}
\begin{subfigure}[t]{0.34\textwidth}
\vspace{0pt}
\caption{Annealing results}
\label{fig:app-path-annealing-table}
\raisebox{-\height}{\usebox{\pathannealingtablebox}}
\end{subfigure}\hfill%
\begin{subfigure}[t]{0.31\textwidth}
\vspace{0pt}
\caption{Validation accuracy}
\label{fig:app-path-annealing-accuracy}
\raisebox{-\height}{\includegraphics[width=\linewidth,
  height=\dimexpr\pathannealingpanelheight+5pt\relax]{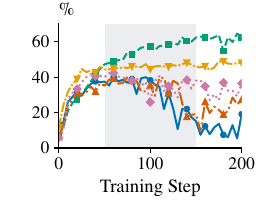}}
\end{subfigure}\hfill%
\begin{subfigure}[t]{0.31\textwidth}
\vspace{0pt}
\caption{Response length}
\label{fig:app-path-annealing-length}
\raisebox{-\height}{\includegraphics[width=\linewidth,
  height=\dimexpr\pathannealingpanelheight+5pt\relax]{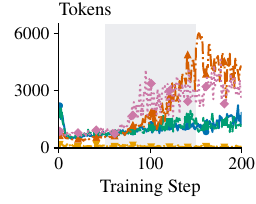}}
\end{subfigure}
\par\vspace{4pt}
\includegraphics[width=\textwidth]{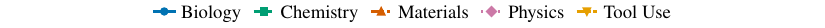}
\captionsetup{skip=6pt}
\caption{Path annealing on Llama-3.2-1B-Instruct~\citep{meta2024llama32}.
(a) Validation Acc/mean@16 percentages at trainer steps 50, 150, and 200.
(b) Validation accuracy every five steps, with higher being better.
(c) Mean training-rollout length in tokens at every step, with no preferred
direction. Shared colors and line styles identify tasks. Shading marks
the linear transition over steps 50--150.}
\label{fig:app-path-annealing}
\end{figure}

\paragraph{Annealing Protocol and Results.}
To test gradual switching, we trained
Llama-3.2-1B-Instruct~\citep{meta2024llama32} for 200 trainer steps per task
with EMA decay 0.95, FEC-FKL, and all-trajectory evidence eligibility.
For trainer step $s$, the progress $a(s)=\clip((s-50)/100,0,1)$ sets
$\lambda_{\mathrm{evi}}=a(s)$ and $\lambda_{\mathrm{adv}}=1-a(s)$.
This transfers weight from advantage modulation to evidence regularization
over steps 50--150, retaining the stopped ZPD controller,
$\lambda_{\mathrm{ref}}=0.1$, and the base
GRPO~\citep{shao2024deepseekmath} loss throughout.
Figure~\ref{fig:app-path-annealing} shows the stage scores and full traces.
Gradual switching does not consistently preserve earlier gains, and the
training dynamics show changes in response length alongside performance.

\paragraph{Interpretation.}
Advantage modulation rescales sampled-action reinforcement, while evidence
regularization changes probability allocation. These different update
targets can favor different optimization trajectories and output behaviors.
One possible explanation is that the two paths induce different effective optimization objectives and could therefore favor different optimal policy distributions. Switching between or jointly activating them could redirect optimization, making earlier gains difficult to preserve or combine.

\subsection{Trajectory-Level Correction Scope}
\label{sec:app-correction-scope}

To examine the role of the evidence mask, Table~\ref{tab:llama-scope-ablation}
compares routing only incorrect trajectories with routing both outcome
classes for each of the three update configurations. All six rows use
Llama-3.2-1B on the same five tasks.
The comparison concerns trajectory eligibility, which precedes token-wise
acceptance. An eligible token still receives no evidence weight when its
local alignment is nonpositive.

\begin{table}[h]
\centering
\small
\caption{Acc/mean@16 by
evidence-correction scope on Llama-3.2-1B.
Wrong-only makes only incorrect trajectories eligible, whereas All makes
both outcome classes eligible. All entries are percentages, with percentage
symbols omitted. Average is the unweighted five-task mean.
Bold marks the higher score within each update-path pair.}
\label{tab:llama-scope-ablation}
\renewcommand{\arraystretch}{1.05}
\setlength{\tabcolsep}{4pt}
\begin{tabular*}{\textwidth}{@{\extracolsep{\fill}}lrrrrrr@{}}
\toprule
Scope & Biology & Chemistry & Materials & Physics & Tool Use & Average \\
\midrule
\multicolumn{7}{@{}l}{\textit{VERPO-LW}} \\
Wrong-only & 56.32 & 64.40 & 44.01 & 48.20 & 45.49 & 51.68 \\
All & \textbf{60.15} & \textbf{66.57} & \textbf{52.26} & \textbf{54.22} & \textbf{49.63} & \textbf{56.57} \\
\midrule
\multicolumn{7}{@{}l}{\textit{VERPO-AM}} \\
Wrong-only & \textbf{57.65} & 63.35 & \textbf{60.30} & 48.67 & 46.37 & \textbf{55.27} \\
All & 55.50 & \textbf{63.69} & 55.06 & \textbf{50.70} & \textbf{51.01} & 55.19 \\
\midrule
\multicolumn{7}{@{}l}{\textit{VERPO-LW+AM}} \\
Wrong-only & 53.66 & 63.51 & 48.40 & \textbf{45.54} & 45.86 & 51.39 \\
All & \textbf{55.38} & \textbf{70.41} & \textbf{56.38} & 44.29 & \textbf{49.26} & \textbf{55.14} \\
\bottomrule
\end{tabular*}
\end{table}

For VERPO-LW, All reaches 0.5657 compared with 0.5168 for Wrong-only
and scores higher on every task. For VERPO-LW+AM, the corresponding averages
are 0.5514 and 0.5139. All leads on four tasks, while Wrong-only leads on Physics.
VERPO-AM instead gives a slightly higher average with Wrong-only,
0.5527 compared with 0.5519. Wrong-only leads on Biology and Materials,
while All leads on Chemistry, Physics, and Tool Use.
Thus the preferred scope depends on the update path and task.
This training comparison differs from the offline analysis below, which
evaluates acceptance on both classes without changing the training mask.

\subsection{Token-Level FEC and ZPD Weight Analysis}
\label{sec:token-weight-analysis}

To locate acceptance within a response, we analyze 100 matched Materials pairs, each containing one format-valid correct and incorrect rollout from the same prompt group. We replay all responses offline without parameter updates and report candidate-level acceptance over both outcome classes, covering 50,292 correct-side and 50,241 incorrect-side token events. 

\paragraph{Trajectory-level Acceptance.}
Table~\ref{tab:fec-token-aggregate} summarizes the aggregate weight
statistics in Figure~\ref{fig:fec-token-analysis-summary}.
Correct rollouts have an effective-weight fraction of 43.3\% compared with
36.4\% for incorrect rollouts, and their mean weights are 0.2337 and 0.2034.
The mean therefore supports broader acceptance in this sample without
implying that every correct rollout receives more weight.

\begin{table}[h]
\centering
\small
\caption{
Token-level FEC and ZPD weight statistics for Llama-3.2-1B. Effective denotes $w>10^{-3}$
and High-weight denotes $w\geq0.5$. Rates and means include zero weights.
The final column reports the
mean trajectory-level sum of controller weights.
}
\label{tab:fec-token-aggregate}
\renewcommand{\arraystretch}{1.08}
\begin{tabular}{lcccc}
\toprule
Trajectory &
Effective \% &
High-weight \% &
Mean Weight &
Mean Sum/Pair \\
\midrule
Correct   & 43.3 & 24.4 & 0.2337 & 117.55 \\
Incorrect & 36.4 & 21.5 & 0.2034 & 102.21 \\
\bottomrule
\end{tabular}
\end{table}

\paragraph{Weight by Response Region.}
We partition each trajectory at its \texttt{<answer>} opening tag.
Tokens before that tag form the reasoning region, and the remaining tokens
form the final-answer region. Table~\ref{tab:fec-token-region} shows that the
mean reasoning-region weight sums are 117.22 and 101.92 for the two classes,
compared with 0.33 and 0.29 in their final answers. Region length contributes
to these totals. The higher per-token means in reasoning show that the
difference also reflects weight allocation, not only the number of tokens.

\begin{table}[h]
\centering
\small
\caption{
FEC and ZPD weight statistics by response region under the same matched replay protocol and thresholds as Table~\ref{tab:fec-token-aggregate}. The
final column reports the mean within-region sum of controller weights.
}
\label{tab:fec-token-region}
\renewcommand{\arraystretch}{1.08}
\begin{tabular}{llcccc}
\toprule
Trajectory & Region &
Effective \% &
High-weight \% &
Mean Weight &
Mean Sum/Pair \\
\midrule
Correct
& Reasoning    & 43.5 & 24.7 & 0.2369 & 117.22 \\
& Final answer & 30.7 & 4.5  & 0.0406 & 0.33 \\
\midrule
Incorrect
& Reasoning    & 36.3 & 21.8 & 0.2062 & 101.92 \\
& Final answer & 40.6 & 3.0  & 0.0359 & 0.29 \\
\bottomrule
\end{tabular}
\end{table}

\paragraph{High-weight Token Types.}
To examine the lexical content of highly weighted positions, we
aggregate weights by token identity. Token types are retained only when
they occur at least 20 times, and their mean weights include
zero-weight occurrences rather than conditioning on controller
acceptance. Table~\ref{tab:fec-token-types} reports high-ranking
lexical or subword types for each trajectory class and provides the
statistics summarized in Figure~\ref{fig:fec-token-analysis-summary}.

\begin{table}[h]
\centering
\small
\caption{
High-ranking token types in the Materials replay sample. Rankings require
at least 20 occurrences and include zero-weight occurrences.
}
\label{tab:fec-token-types}
\renewcommand{\arraystretch}{1.04}
\begin{tabular}{llrrrr}
\toprule
Trajectory & Token & Occurrences &
Effective \% & High-weight \% & Mean Weight \\
\midrule
Correct
& \texttt{analyze}   & 23  & 73.9 & 73.9 & 0.685 \\
& \texttt{Given}     & 61  & 68.9 & 67.2 & 0.614 \\
& \texttt{select}    & 25  & 68.0 & 64.0 & 0.612 \\
& \texttt{therm}     & 23  & 65.2 & 65.2 & 0.607 \\
& \texttt{compound}  & 20  & 80.0 & 65.0 & 0.606 \\
& \texttt{greater}   & 30  & 66.7 & 63.3 & 0.597 \\
& \texttt{one}       & 30  & 73.3 & 60.0 & 0.596 \\
& \texttt{reference} & 140 & 64.3 & 62.9 & 0.589 \\
\midrule
Incorrect
& \texttt{without}   & 24 & 62.5 & 62.5 & 0.579 \\
& \texttt{compute}   & 24 & 62.5 & 58.3 & 0.547 \\
& \texttt{indicates} & 35 & 60.0 & 57.1 & 0.545 \\
& \texttt{actually}  & 20 & 60.0 & 60.0 & 0.540 \\
& \texttt{these}     & 54 & 64.8 & 55.6 & 0.523 \\
& \texttt{Therefore} & 34 & 67.6 & 52.9 & 0.515 \\
& \texttt{find}      & 49 & 59.2 & 57.1 & 0.515 \\
& \texttt{will}      & 28 & 53.6 & 53.6 & 0.506 \\
\bottomrule
\end{tabular}
\end{table}

The correct-side ranking includes \texttt{analyze}, \texttt{select}, and
\texttt{reference}, alongside domain-related types such as \texttt{therm}.
The incorrect-side ranking includes \texttt{compute}, \texttt{without},
and \texttt{Therefore}. These examples describe lexical associations with
acceptance. A token-type mean merges different contexts and does not identify
a causal reasoning operation or an error location.

\paragraph{Matched Token Types.}

The previous ranking can partly reflect differences in which words are
used by the two classes. To make the comparison more direct, we next
restrict attention to token types occurring at least 20 times in
\emph{both} classes and compare their mean weights. Table~
\ref{tab:fec-token-differences} reports the largest differences in each
direction.

\begin{table}[h]
\centering
\small
\caption{
Largest differences in mean controller weight between correct and incorrect
rollouts among
token types occurring at least 20 times in both classes. For the left
pair of columns, the gap is $\bar{w}_{\mathrm{correct}}-
\bar{w}_{\mathrm{incorrect}}$. For the right pair,
the gap is $\bar{w}_{\mathrm{incorrect}}-
\bar{w}_{\mathrm{correct}}$.
}
\label{tab:fec-token-differences}
\renewcommand{\arraystretch}{1.05}
\begin{tabular}{lrrlrr}
\toprule
\multicolumn{3}{c}{Correct-favored} &
\multicolumn{3}{c}{Incorrect-favored} \\
\cmidrule(lr){1-3}\cmidrule(lr){4-6}
Token & Correct & Gap &
Token & Incorrect & Gap \\
\midrule
\texttt{therm}      & 0.607 & +0.370 &
\texttt{without}    & 0.579 & +0.234 \\
\texttt{Given}      & 0.614 & +0.360 &
\texttt{Correct}    & 0.492 & +0.226 \\
\texttt{greater}    & 0.597 & +0.327 &
\texttt{P}          & 0.321 & +0.217 \\
\texttt{sqrt}       & 0.527 & +0.305 &
\texttt{Therefore}  & 0.515 & +0.216 \\
\texttt{incorrect}  & 0.520 & +0.290 &
\texttt{indicates}  & 0.545 & +0.180 \\
\texttt{element}    & 0.525 & +0.281 &
\texttt{does}       & 0.466 & +0.178 \\
\texttt{reference}  & 0.589 & +0.221 &
\texttt{properties} & 0.459 & +0.177 \\
\bottomrule
\end{tabular}
\end{table}

For example, \texttt{therm} has a correct-favored gap of 0.370, while
\texttt{without} has an incorrect-favored gap of 0.234. Comparing the same
token types removes differences in type identity but does not match their
surrounding contexts. 

\subsection{Hyperparameter Sensitivity}
\label{sec:hyperparameter_sensitivity}

To examine the evidence-loss scale in Equation~\eqref{eq:verpo-objective},
we extend Table~\ref{tab:lambda-evi-ablation} with the first 200 training
steps of five Llama-3.2-1B Chemistry runs.
The four settings $\lambda_{\mathrm{evi}}\in\{0.2,0.4,0.6,0.8\}$ used
all-trajectory evidence correction. All five used FEC, FKL, EMA decay
0.95, $\lambda_{\mathrm{ref}}=0.1$, and loss weighting without advantage
modulation.

\begin{figure}[h]
    \centering
    \includegraphics[width=\textwidth]{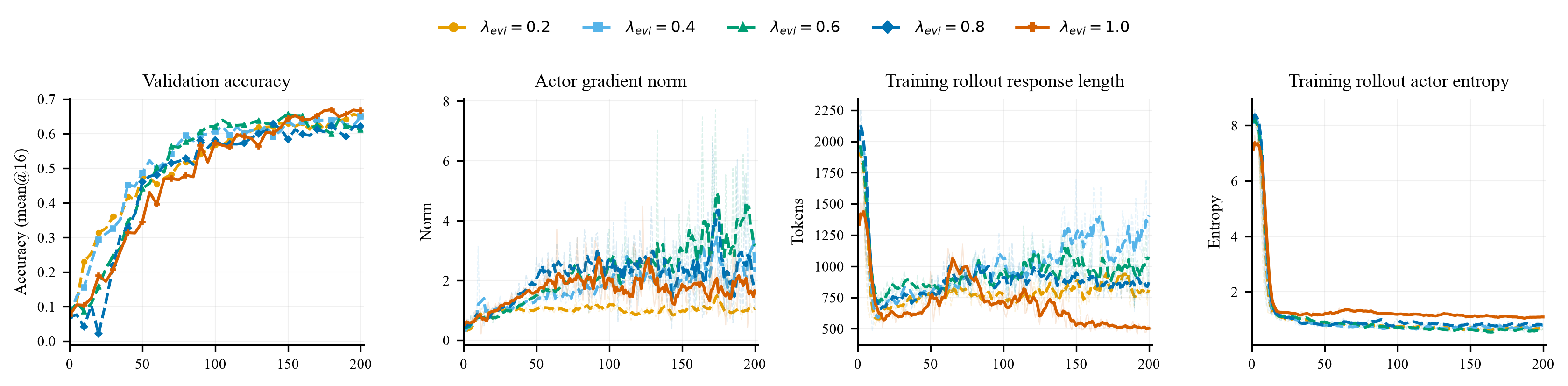}
    \caption{
        Sensitivity of Llama-3.2-1B to the evidence-loss coefficient
        $\lambda_{\mathrm{evi}}$ on Chemistry.
        From left to right, the panels report validation accuracy, Actor
        gradient norm, rollout response length, and Actor entropy.
        Higher accuracy indicates better task performance. 
    }
    \label{fig:llama-chemistry-lambda-sensitivity}
\end{figure}

Figure~\ref{fig:llama-chemistry-lambda-sensitivity} and
Table~\ref{tab:lambda-evi-ablation} show no monotonic improvement with
$\lambda_{\mathrm{evi}}$.
The 1.0 run reaches the highest step-200 score of 66.58\%, but its
code revision and planned horizon differ. This comparison cannot attribute
that advantage to $\lambda_{\mathrm{evi}}$ alone or establish 1.0 as the
best coefficient under a matched protocol.

\subsection{Probability Mass Retention under Top-K Truncation}
\label{sec:app-topk-support}

To assess the vocabulary approximation in
Appendix~\ref{sec:app-method-hyperparameters}, we examine the first 200 steps
of five Llama-3.2-1B-Instruct~\citep{meta2024llama32} FEC runs, one per task.
All use FKL, joint loss weighting and advantage modulation, all-trajectory
evidence routing, and EMA decay 0.95. 

\begin{figure}[!htb]
    \centering
    \includegraphics[width=\textwidth]{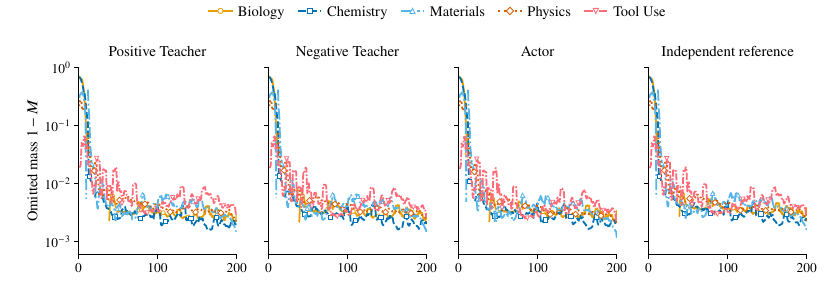}
    \caption{Omitted probability mass over the first 200 training steps for
    five FEC-LW+AM runs on Llama-3.2-1B-Instruct~\citep{meta2024llama32}.
    The underlying observations are step means over routed valid response tokens.
    Lower values indicate greater probability retention on the selected
    support. The first three panels use the shared evidence-correction
    support, and the reference panel uses its independent support.
    The logarithmic axes show centered five-step moving averages within
    contiguous valid segments, with segment endpoints preserved.
    Colors, line styles, and markers distinguish tasks. Gaps mark zero-token
    steps and are not interpolated.}
    \label{fig:topk-support-mass}
\end{figure}

\paragraph{Support Construction and Measurement.}
The correction support $S$ combines the top 128 candidates from each of
$q^+$, $q^-$, and $q^0$ with the sampled token and removes duplicates.
Thus $K=128$ specifies candidates per branch, not the final support size.
The reference channel uses a separate support $S_{\mathrm{ref}}$ formed
from its own top 128 candidates and the sampled token.
All selected probabilities retain the full softmax normalization, without
renormalizing the selected coordinates or adding a tail bucket.
For a distribution $q$, retained mass on $S$ is
$M=\sum_{v\in S}q(v)$.
Figure~\ref{fig:topk-support-mass} reports the logged token means of $1-M$.
The positive Teacher, negative Teacher, and Actor share $S$, whereas the
reference mass uses $S_{\mathrm{ref}}$.

\paragraph{Observed Retention.}
All numerical summaries use unsmoothed observations. Each logged mass
averages over routed valid response tokens before acceptance weighting.
We aggregate steps using their logged token counts,
rather than averaging step means equally. 
Across the five tasks and four branches, token-weighted retained mass ranges
from 95.92\% to 99.32\%, while the step-200 means range from 99.62\% to
99.88\%. The task-specific token-weighted mean correction-support sizes
range from 148.26 to 171.33 coordinates.
These measurements support $K=128$ as a compact support with high aggregate
probability retention in the tested configuration.
Retention is not uniformly high throughout training. The lowest valid step
mean is 28.90\% in the early Chemistry reference trace, and the plot retains
these early low-coverage observations.

\paragraph{Exact Truncated Gradient.}
To identify what truncation changes, consider one token with
$p=\operatorname{softmax}(z)$ at unit temperature. Freeze $S$, the signed
direction $\Delta$ on its selected coordinates, and the nonnegative weight
$w$ during differentiation. Define the zero-padded vector
$\Delta_S(v)=\mathbf{1}[v\in S]\Delta(v)$ and the truncated loss
$L_S=-w\sum_{v\in S}\Delta(v)\log p(v)$.
The softmax derivative
$\partial\log p(v)/\partial z_j=\mathbf{1}[v=j]-p(j)$ gives
\begin{equation}
s_S=\sum_{v\in S}\Delta(v),
\qquad
-\nabla_z L_S=w\bigl(\Delta_S-s_Sp\bigr).
\label{eq:app-topk-gradient}
\end{equation}
The extra term $-ws_Sp$ comes from the live full-vocabulary normalizer.
It acts on selected and omitted logits, ensuring that the logit update sums
to zero even when the selected direction does not.
Since $\|p\|_1=1$ and $w\geq0$, its absolute size satisfies
\begin{equation}
\bigl\|-\nabla_z L_S-w\Delta_S\bigr\|_1=w|s_S|.
\label{eq:app-topk-gradient-error}
\end{equation}
Thus the simple direction $w\Delta_S$ recovers the negative loss gradient when
$s_S=0$, including a full-vocabulary difference of normalized endpoints.
A small-vocabulary finite-difference check verifies both identities for
nonzero mass defect, zero mass defect, full support, and zero weight, with
maximum coordinate error below $3.2\times10^{-12}$.

\section{Diagnostic Metrics}
\label{sec:app-diagnostics}

To interpret the controller traces in Figure~\ref{fig:fec-diagnostics}, this
appendix defines their token sets, averaging rules, and inference limits.
The four diagnostics describe acceptance and the measured FEC residual during
training. They are not validation or test metrics.
Let $\mathcal I_{\mathrm{route}}$ denote the valid
response-token positions included by the evidence-route diagnostic mask. The
implementation first sums each statistic and its token count across
microbatches and data-parallel workers and then divides. Every mean below is
therefore a global token-weighted mean rather than a mean of per-batch means.
For the displayed FEC runs, the FEC diagnostic set equals
$\mathcal I_{\mathrm{route}}$. The mean definitions below require a nonempty
diagnostic set. They do not specify a logging convention for empty sets.

\subsection{ZPD Effective Weight Coverage}
\label{sec:diagnostic-zpd}

To measure the breadth of acceptance, $\mathrm{WeightEffectiveCoverage}$
counts diagnostic tokens whose stopped weight exceeds a fixed monitoring
threshold:
\begin{equation}
\mathrm{WeightEffectiveCoverage}
=
\frac{1}{\left|\mathcal I_{\mathrm{route}}\right|}
\sum_{(i,t)\in\mathcal I_{\mathrm{route}}}
\mathbf{1}\!\left[w_{i,t}>\tau_{\mathrm{eff}}\right],
\qquad
\tau_{\mathrm{eff}}=10^{-3}.
\label{eq:diagnostic-weight-effective-coverage}
\end{equation}
The denominator counts routed diagnostic tokens, not every token in the
batch. Coverage therefore measures acceptance conditional on this set and
does not measure how often the routing mask selects tokens.
The threshold is used only for
monitoring and does not enter the controller. This metric differs from the
strict nonzero fraction $\mathbf 1[w_{i,t}>0]$, which would count arbitrarily
small numerical weights as active.

\subsection{Mean Signed Benefit and Fisher Cost}
\label{sec:diagnostic-benefit-cost}

To distinguish favorable and conflicting directions before rejection, we
retain signed alignment in the benefit diagnostic.
Equation~\eqref{eq:zpd-proxies-main} applies a positive part to this value
for controller use. To expose rejected as well as accepted
directions, the logged benefit diagnostic retains the value before this
positive-part operation:
\begin{equation}
b_{i,t}
=\widehat A_i(e_{y_{i,t}}-p_{i,t})^\top u_{i,t},
\qquad
h_{i,t}=[b_{i,t}]_+.
\label{eq:diagnostic-raw-benefit}
\end{equation}
The one-hot score direction compares the evidence move with the sampled
token's outcome signal. Only $h_{i,t}$ enters the default acceptance rule.
The logged metric $\mathrm{BenefitMean}$, shown in the second panel of
Figure~\ref{fig:fec-diagnostics}, is
\begin{equation}
\mathrm{BenefitMean}
=\frac{1}{|\mathcal I_{\mathrm{route}}|}
\sum_{(i,t)\in\mathcal I_{\mathrm{route}}}b_{i,t}.
\label{eq:diagnostic-benefit-mean}
\end{equation}
It is a raw signed mean, not the mean of $h_{i,t}$ and not a conditional mean
over positive-benefit tokens. A negative value means that the signed evidence
directions oppose the local GRPO~\citep{shao2024deepseekmath} score direction
on average. The default controller assigns zero weight to each token with
nonpositive benefit, but a negative mean does not imply that every token is
rejected. It also does not measure the reward change after an optimizer step.

The logged metric $\mathrm{FisherCostMean}$, shown in the third panel,
is the mean nonnegative Fisher movement term from
Equation~\eqref{eq:zpd-proxies-main}:
\begin{equation}
\mathrm{FisherCostMean}
=\frac{1}{|\mathcal I_{\mathrm{route}}|}
\sum_{(i,t)\in\mathcal I_{\mathrm{route}}}c_{i,t},
\qquad
c_{i,t}=u_{i,t}^{\top}F(p_{i,t})u_{i,t}\geq0.
\label{eq:diagnostic-fisher-cost-mean}
\end{equation}
This nonnegative statistic measures local distribution movement, not
observed reward loss. It is the unscaled dynamic term in the controller denominator. The full
movement penalty is
$\alpha_{\mathrm{cost}}c_{i,t}+\epsilon_{\mathrm{cost}}$, so the plotted
quantity excludes both the scale and the positive floor. 

\subsection{FEC Residual Nuisance Fisher Covariance}
\label{sec:diagnostic-fec}

To check the aggregate residual left by the projection,
$\mathrm{FECResidualCov}$ averages the signed Fisher covariance
between the retained FEC residual in Equation~\eqref{eq:main-fec-direction}
and the measured nuisance direction in
Equation~\eqref{eq:main-fec-coefficient}:
\begin{equation}
\mathrm{FECResidualCov}
=
\frac{1}{\left|\mathcal I_{\mathrm{FEC}}\right|}
\sum_{(i,t)\in\mathcal I_{\mathrm{FEC}}}
\left\langle
\Delta_{i,t}^{\mathrm{FEC}},
\Delta_{i,t}^{\mathrm{nuis}}
\right\rangle_{F(p_{i,t})}.
\label{eq:diagnostic-fec-residual-covariance}
\end{equation}
Here $\mathcal I_{\mathrm{FEC}}$ contains the valid response-token positions
on which the FEC branches and projection diagnostics are available.
Equation~\eqref{eq:app-fec-fkl-residual} gives the remaining ridge-dependent
covariance for each token under the coefficient in
Equation~\eqref{eq:main-fec-coefficient}.

A small mean signed covariance does not establish token-wise orthogonality.
Positive and negative residuals can cancel in the average, and small vector
norms also produce small covariance without a small angle.
Figure~\ref{fig:fec-diagnostics} therefore provides an aggregate numerical
check of the projection rather than a bound on every residual.
Neither the sign nor the magnitude establishes a causal contribution to
task accuracy.

\section{Experimental Details}
\label{sec:app-details}

To document the comparisons in Section~\ref{sec:experiments}, this appendix
collects the data splits, runtime, shared training settings, method-specific
coefficients, and supplied prompt templates. Hardware-specific timing runs
are reported separately in Appendix~\ref{sec:computational-analysis}.

\subsection{Datasets and Data Processing}
\label{sec:app-datasets}

We use the five-task benchmark release associated with SRPO~\citep{li2026srpo}
as a fixed data source. Biology, Chemistry, Materials Science, and Physics are
the Level-3 reasoning subsets of
SciKnowEval~\citep{feng2024sciknoweval}. Tool Use is derived from
ToolAlpaca~\citep{tang2023toolalpaca}. Table~\ref{tab:benchmark-data} reports
the complete train and test splits used by every compared method. 

\begin{table}[H]
\centering
\small
\caption{Five-task data splits shared by the compared methods. Counts denote
examples rather than sampled responses.}
\label{tab:benchmark-data}
\renewcommand{\arraystretch}{1.08}
\begin{tabular}{llrrr}
\toprule
Task & Source & Train & Test & Total \\
\midrule
Biology & SciKnowEval~\citep{feng2024sciknoweval} Level 3 & 450 & 50 & 500 \\
Chemistry & SciKnowEval~\citep{feng2024sciknoweval} Level 3 & 1,890 & 210 & 2,100 \\
Materials Science & SciKnowEval~\citep{feng2024sciknoweval} Level 3 & 841 & 94 & 935 \\
Physics & SciKnowEval~\citep{feng2024sciknoweval} Level 3 & 720 & 80 & 800 \\
Tool Use & ToolAlpaca~\citep{tang2023toolalpaca} & 4,046 & 68 & 4,114 \\
\midrule
Total & -- & 7,947 & 502 & 8,449 \\
\bottomrule
\end{tabular}
\end{table}
The fixed splits prevent data membership from changing across methods.
Science prompts request one of four answer options, while Tool Use prompts
provide the available tool specifications. The literal input scaffolds are
given in Appendix~\ref{sec:app-prompts}. For each trained method, backbone, and task, Table~\ref{tab:main-results}
reports the highest evaluation score observed during training. Base rows
report the initial model scores. Average is the unweighted mean of the five
task-specific maxima, not performance at a common or final checkpoint.
For a collapsed run, the reported maximum is its pre-collapse peak and is
marked with $\dagger$. Specifically, the Llama-3.2-1B SDPO~\citep{hubotter2026sdpo} Physics score of 40.15
is the highest pre-collapse score from the run in
Figure~\ref{fig:training-dynamics-physics}. The displayed curve peaks at
step 100 and reaches zero at steps 160, 180, and 200.
The score and its five-task average are marked with $\dagger$ and excluded
from comparisons restricted to complete, noncollapsed runs.
This average is descriptive and does not represent final performance.
Table~\ref{tab:lambda-evi-ablation} instead reports fixed step-200 scores
and late-window summaries. Marked entries in
Table~\ref{tab:uniform-weight-ablation} report pre-collapse peaks.

\subsection{Models, Baselines, and Runtime}
\label{sec:app-technical-setup}

To separate model and runtime choices from method coefficients, we summarize
the Actor initialization, hardware profile, and
thinking-mode switches are summarized in Table~\ref{tab:training-protocol}.
The two-GPU Llama cost study in
Appendix~\ref{sec:computational-analysis} is a separate hardware setting and
is not pooled with the eight-GPU timing comparison.
The implementation is built on \texttt{verl}~\citep{sheng2024hybridflow}, uses
PyTorch FSDP2~\citep{feng2025fsdp2tutorial} for distributed Actor training, and
uses vLLM~\citep{kwon2023efficient} for batched rollout generation. We use
AdamW~\citep{loshchilov2019decoupled} for optimization.

\subsection{Shared Training Setup}
\label{sec:app-shared-setup}

To define the common sampling and optimization budget,
Table~\ref{tab:training-protocol} separates rollout settings from validation
decoding. These settings describe matched benchmark runs, not the
hardware-specific cost profiles. Table~\ref{tab:verpo-hyperparameters}
then isolates the coefficients and support approximation used by VERPO.

\begin{table}[H]
\centering
\small
\caption{Common benchmark, runtime, and optimization settings used by matched
experiments. }
\label{tab:training-protocol}
\renewcommand{\arraystretch}{1.08}
\begin{tabularx}{\textwidth}{@{}lY@{}}
\toprule
Parameter & Value \\
\midrule
Actor initialization & Qwen3-4B~\citep{yang2025qwen3}, Qwen3-8B~\citep{yang2025qwen3},
Llama-3.2-1B-Instruct~\citep{meta2024llama32} \\
Hardware & 8$\times$ NVIDIA A800 GPUs per run \\
Actor / Teacher / validation thinking & False / False / False \\
Global prompt batch / rollouts per prompt & 32 / 8 \\
Mini-batch trajectories / update epochs & 32 / 1 \\
Maximum prompt / response length & 2,048 / 8,192 tokens \\
Maximum reprompt / model length & 10,240 / 18,944 tokens \\
Training decoding & temperature 1.0, top-$p$ 1.0, top-$k$ disabled \\
Validation decoding & 16 rollouts, temperature 0.6, top-$p$ 0.95, top-$k$ disabled \\
Outcome clipping & upper radius $\epsilon_{\mathrm{high}}=0.28$ \\
Optimizer & AdamW~\citep{loshchilov2019decoupled}, weight decay 0.01, gradient clip 1.0 \\
Schedule & constant after 10 warmup steps \\
Budget & 200 optimizer steps \\
\bottomrule
\end{tabularx}
\end{table}

\subsection{Method-Specific Hyperparameters}
\label{sec:app-method-hyperparameters}

To distinguish algorithm settings from the shared budget,
Table~\ref{tab:baseline-protocols} lists the GRPO, SDPO, and SRPO
configurations, while Table~\ref{tab:rlsd-rlcsd-hyperparameters} reports the
RLSD and RLCSD reproduction settings. The SDPO and SRPO entries use the
symmetric Jensen-Shannon divergence~\citep{lin1991divergence}, distinct from
the FKL and RKL decompositions used for VERPO.
Table~\ref{tab:verpo-hyperparameters} separates VERPO's reference and evidence
coefficients from its controller parameters.
Given the benefit and cost proxies, the acceptance fraction uses
$\alpha_{\mathrm{cost}}$ and $\epsilon_{\mathrm{cost}}$.
The projection ridge instead changes the evidence direction before weighting.

\begin{table}[H]
\centering
\small
\caption{VERPO-specific hyperparameters.}
\label{tab:verpo-hyperparameters}
\renewcommand{\arraystretch}{1.08}
\begin{tabular}{@{}ll@{}}
\toprule
Parameter & Value \\
\midrule
$\lambda_{\mathrm{ref}}$ & 0.1 \\
$\lambda_{\mathrm{evi}}$ & 1.0 \\
Direction construction & Fix~\citep{hubotter2026sdpo}, CTR~\citep{pan2026rlcsd}, FEC \\
KL geometry & FKL, RKL \\
ZPD cost coefficient $\alpha_{\mathrm{cost}}$ & 0.0025 \\
ZPD denominator floor $\epsilon_{\mathrm{cost}}$ & $2.5\times10^{-5}$ \\
FEC ridge & $10^{-8}$ \\
Teacher candidates per branch & Top-$K=128$ \\
Vocabulary chunk size & 4,096  \\
EMA Teacher decay & 0.95 \\
\bottomrule
\end{tabular}
\end{table}

\begin{table}[H]
\centering
\small
\caption{Baseline-specific loss and optimizer settings. Dashes preserve
entries not specified for the corresponding baseline.}
\label{tab:baseline-protocols}
\renewcommand{\arraystretch}{1.06}
\begin{tabularx}{\textwidth}{@{}lYYY@{}}
\toprule
Parameters & GRPO~\citep{shao2024deepseekmath} &
SDPO~\citep{hubotter2026sdpo} & SRPO~\citep{li2026srpo} \\
\midrule
\multicolumn{4}{@{}l@{}}{\textbf{Outcome loss}} \\

\midrule

Asymmetric upper clip $\epsilon_{\mathrm{high}}$ & 0.28 & -- & 0.28 \\
Rollout IS clip $\rho$ & -- & -- & 2.0 \\
KL coefficient & 0.0 & -- & 0.0 \\

\midrule

\addlinespace
\multicolumn{4}{@{}l@{}}{\textbf{Distillation loss}} \\

\midrule

Top-$K$ distillation & -- & 100  & 100  \\
Distillation divergence & -- & Jensen-Shannon~\citep{lin1991divergence} & Jensen-Shannon~\citep{lin1991divergence} \\
EMA update rate & -- & 0.05 & 0.05 \\
Rollout IS clip $\rho$ & -- & 2.0 & 2.0 \\

\midrule

\addlinespace
\multicolumn{4}{@{}l@{}}{\textbf{Dynamic weighting}} \\

\midrule

Dynamic-weighting coefficient $\beta$ & -- & -- & 1 \\

\midrule

\addlinespace
\multicolumn{4}{@{}l@{}}{\textbf{Training}} \\

\midrule

Optimizer & AdamW~\citep{loshchilov2019decoupled} &
AdamW~\citep{loshchilov2019decoupled} & AdamW~\citep{loshchilov2019decoupled} \\
Learning rate & $5\times10^{-6}$ & $1\times10^{-5}$ & $5\times10^{-6}$ \\
Warmup steps & 10 & 10 & 10 \\
Weight decay & 0.01 & 0.01 & 0.01 \\
Gradient clip norm & 1.0 & 1.0 & 1.0 \\
\bottomrule
\end{tabularx}
\end{table}

\begin{table}[H]
\centering
\small
\caption{RLSD and RLCSD reproduction hyperparameters in our unified
PGR-Probe baseline implementation. The method references and official
repositories are RLSD~\citep{yang2026rlsd}
(\href{https://github.com/iie-ycx/RLSD}{code}) and
RLCSD~\citep{pan2026rlcsd}
(\href{https://github.com/THU-BPM/RLCSD}{code}). Dashes preserve entries not
used by the corresponding baseline.}
\label{tab:rlsd-rlcsd-hyperparameters}
\renewcommand{\arraystretch}{1.06}
\begin{tabularx}{\textwidth}{@{}lYY@{}}
\toprule
Parameters & RLSD~\citep{yang2026rlsd} & RLCSD~\citep{pan2026rlcsd} \\
\midrule
\multicolumn{3}{@{}l@{}}{\textbf{Outcome loss}} \\

\midrule

Asymmetric upper clip $\epsilon_{\mathrm{high}}$ & 0.28 & 0.28 \\
Rollout IS clip $\rho$ & 2.0 & 2.0 \\
KL coefficient & 0.0 & 0.0 \\

\midrule

\addlinespace
\multicolumn{3}{@{}l@{}}{\textbf{Teacher construction}} \\

\midrule

Teacher mode / EMA update rate & EMA / 0.05 & EMA / 0.05 \\
Teacher temperature & 1.0 & 1.0 \\
Maximum negative hints & -- & 4 \\
Remove thinking from demonstration & True & True \\

\midrule

\addlinespace
\multicolumn{3}{@{}l@{}}{\textbf{Advantage modulation}} \\

\midrule

Modulation coefficient $\lambda$ & 0.5 & 0.5 \\
Evidence-ratio clip & 0.2 & -- \\
Linear decay steps & 50 & -- \\
Contrast temperature $\tau$ & -- & 0.02 \\
Contrast scale $\beta$ & -- & 1.0 \\
Selection threshold $\delta$ & -- & 0.02 \\
Residual clip & -- & $[-2,2]$ \\
Selected-path weight & -- & 1.0 \\

\midrule

\addlinespace
\multicolumn{3}{@{}l@{}}{\textbf{Training}} \\

\midrule

Optimizer & AdamW~\citep{loshchilov2019decoupled} &
AdamW~\citep{loshchilov2019decoupled} \\
Learning rate & $5\times10^{-6}$ & $5\times10^{-6}$ \\
Warmup steps & 10 & 10 \\
Weight decay & 0.01 & 0.01 \\
Gradient clip norm & 1.0 & 1.0 \\
\bottomrule
\end{tabularx}
\end{table}

To reduce vocabulary computation, the implementation forms a deduplicated
union of the top-$K$ candidates from the Teacher branches used by each
correction and adds the sampled token. FKL FEC uses the three branches
$q^+$, $q^-$, and $q^0$. The RKL correction additionally includes the Actor's
top-$K$ candidates. Thus $K=128$ specifies candidates per branch, not the
size of the final union.

Selected log probabilities retain their full-vocabulary softmax
normalization. The omitted tail is neither renormalized into the selected
coordinates nor represented by an extra bucket. Evidence losses, Fisher
moments, and projection statistics on this support are therefore
approximations to the full-vocabulary quantities in
Appendix~\ref{sec:theoretical-analysis}. In particular, a truncated FKL
displacement need not sum to zero on the selected coordinates, so the exact
full-vocabulary logit identities cannot be assumed unchanged after truncation.
Support selection, Teacher values, and projection coefficients are stopped
during Actor differentiation, while the Actor softmax normalizer remains live.
Appendix~\ref{sec:app-topk-support} gives the exact truncated logit gradient
and measures probability retention for $K=128$ over 200 training steps.

\subsection{Prompt Templates}
\label{sec:app-prompts}

To make the task formatting explicit, the following listings preserve the
supplied prompt text, including literal placeholders, punctuation, and
response tags. The science system message constrains the final output to
an option letter, while its user message supplies the question and options.

\begin{promptbox}[lst:science-system]{Science system message}
Given a question and four options, please select the right answer. Respond in the following format:
<reasoning>
...
</reasoning>
<answer>
...
</answer>

For the answer, only output the letter corresponding to the correct option (A, B, C, or D), and nothing else. Do not restate the answer text. For example, if the answer is "A", just output:
<answer>
A
</answer>
\end{promptbox}

\begin{promptbox}[lst:science-user]{Science user message}
[record-specific question and four answer options]
Please reason step by step.
\end{promptbox}

\noindent
The Tool Use messages restrict the Actor to the supplied tools and specify
the action-call format. Record-specific API descriptions and schemas occupy
the indicated placeholder rather than a shared fixed tool list.

\begin{promptbox}[lst:tool-system]{Tool Use system message}
You are a tool-use assistant. Solve each request by reasoning about the task and calling the provided tools when needed.
Use only the tools provided in the user message.
Follow the required response format exactly.
\end{promptbox}

\begin{promptbox}[lst:tool-user]{Tool Use user-message scaffold}
Your task is to answer the user's question using available tools.
You have access to the following tools:
[record-specific tool names, descriptions, API documentation,
 parameter schemas, and output schemas]

Use the following format:
Thought: you should always think about what to do
Action: the action to take, should be one of the tool names.
Action Input: the input to the action, must be in JSON format. All of the action input must be realistic and from the user.

Begin!
Question: [record-specific question]
\end{promptbox}

The final listing shows the positive-evidence Teacher scaffold. It augments
the task prompt with privileged context before replaying the sampled prefix
as defined in Equation~\eqref{eq:app-teacher-replay}.

\begin{promptbox}[lst:self-teacher]{Evidence-conditioned self-Teacher reprompt}
[original task user prompt]
Correct solution:

[validated privileged context or selected sibling response]


Correctly solve the original question.
\end{promptbox}

\section{Computational Analysis}
\label{sec:computational-analysis}

To put the accuracy comparisons in computational context, this appendix
reports logged step costs and the approximate compute-matched comparison
discussed with Table~\ref{tab:main-results}. The Llama-3.2-1B measurements
use two A800 GPUs, while the Qwen3-4B Chemistry measurements use eight.
These hardware groups are evaluated separately.

\subsection{Per-Step Cost}
\label{sec:per-step-cost}

To distinguish total step cost from the recorded component timers,
Tables~\ref{tab:llama-computation-cost} and~\ref{tab:qwen3-4b-computation-cost}
retain the generation, Teacher/evidence, and Actor-update columns alongside
the logged total. Each entry averages its corresponding logged statistic
over steps 10 to 200. 

\begin{table}[h]
\centering
\footnotesize
\setlength{\tabcolsep}{3.5pt}
\caption{Logged Llama-3.2-1B computation costs on two A800 GPUs over
steps 10 to 200. Gen denotes generation. Peak alloc.\ denotes logged peak
allocated memory. }
\label{tab:llama-computation-cost}
\begin{tabularx}{\textwidth}{@{}Y l r r r r r r@{}}
\toprule
Method & Hardware & Gen & Teacher/evidence & Actor update & Total step & Throughput & Peak alloc. \\
 &  & s/step & s/step & s/step & s/step & tok/s/GPU & GB \\
\midrule
Fixed~\citep{hubotter2026sdpo} & 2×A800 & 7.21 & 53.48 & 12.30 & 23.30 & 508.4 & 10.77 \\
CTR~\citep{pan2026rlcsd} & 2×A800 & 10.92 & 53.99 & 14.84 & 29.68 & 638.9 & 11.66 \\
FEC & 2×A800 & 15.53 & 54.20 & 15.58 & 35.20 & 1021.4 & 12.30 \\
\bottomrule
\end{tabularx}
\end{table}

\begin{table}[h]
\centering
\footnotesize
\setlength{\tabcolsep}{3.5pt}
\caption{Logged Qwen3-4B Chemistry computation costs on eight A800 GPUs over
steps 10 to 200. N/A means that the corresponding Teacher/evidence timer
was not available, not that the method performed no Teacher computation.}
\label{tab:qwen3-4b-computation-cost}
\begin{tabularx}{\textwidth}{@{}Y l r r r r r r@{}}
\toprule
Method & Hardware & Gen & Teacher/evidence & Actor update & Total step & Throughput & Peak alloc. \\
 &  & s/step & s/step & s/step & s/step & tok/s/GPU & GB \\
\midrule
GRPO~\citep{shao2024deepseekmath} & 8×A800 & 22.52 & N/A & 17.58 & 48.50 & 466.8 & 16.20 \\
SDPO~\citep{hubotter2026sdpo} & 8×A800 & 3.32 & N/A & 28.84 & 40.80 & 278.8 & 35.07 \\
SRPO~\citep{li2026srpo} & 8×A800 & 5.98 & N/A & 27.51 & 42.20 & 309.2 & 23.07 \\
VERPO & 8×A800 & 31.09 & 176.34 & 42.14 & 82.41 & 281.5 & 18.81 \\
\bottomrule
\end{tabularx}
\end{table}

In Table~\ref{tab:llama-computation-cost},
Fixed~\citep{hubotter2026sdpo} has a logged total of 23.30 seconds per step,
compared with 29.68 for CTR~\citep{pan2026rlcsd} and 35.20 for FEC.
This ranks the observed step costs but does not establish equal training
quality. In particular, the Fixed-direction experiments exhibited reward
hacking as reported with Table~\ref{tab:component-ablation}.
In Table~\ref{tab:qwen3-4b-computation-cost}, VERPO records 82.41 seconds
per step compared with 48.50 for GRPO~\citep{shao2024deepseekmath}.
Its reported peak allocation is 18.81 GB, compared with 16.20 GB for that
baseline and 35.07 GB for SDPO~\citep{hubotter2026sdpo}.
These are measurements from the listed profiles. Throughput alone does not
rank end-to-end efficiency because generated token counts and response
lengths differ. The recorded peak-allocation field also does not establish
whether worker aggregation uses a maximum or another reduction.

\subsection{Compute-Matched Performance}
\label{sec:compute-matched-performance}

To compare quality at an approximately shared budget, we evaluate VERPO,
GRPO~\citep{shao2024deepseekmath}, SDPO~\citep{hubotter2026sdpo}, and
SRPO~\citep{li2026srpo} on the same Llama-3.2-1B backbone and
$2\times$A800 hardware in Table~\ref{tab:llama-same-compute}.
For each run, we estimate the training
compute based on the average wall-clock time per optimization step after the
initial warm-up period. The budget for each task is defined by the compute
required for the corresponding 200-step outcome-only baseline, and the nearest available
five-step validation checkpoint is selected for the other methods. VERPO has the highest reported accuracy in each of the five task columns.
Its average of 0.5026 exceeds the 0.4791 from
GRPO~\citep{shao2024deepseekmath} by 0.0235, at an average of 94 rather
than 200 optimizer steps. 

\begin{table}[h]
\centering
\scriptsize
\setlength{\tabcolsep}{3.2pt}
\renewcommand{\arraystretch}{1.08}
\caption{Approximate compute-matched Llama-3.2-1B results. Each method cell
reports matched step / validation accuracy. The budget row gives the
task-specific target in A800 GPU-hours. }
\label{tab:llama-same-compute}
\begin{tabular*}{\textwidth}{@{\extracolsep{\fill}}lrrrrrr@{}}
\toprule
Method & Biology & Chemistry & Materials & Physics & Tool Use & Average \\
\midrule
Budget, GPU-h & 10.91 & 11.53 & 10.61 & 9.97 & 10.08 & 10.62 \\
\midrule
VERPO & 105 / 51.00 & 105 / 56.61 & 85 / 45.70 & 85 / 49.14 & 90 / 48.85 & 94 / 50.26\\
GRPO~\citep{shao2024deepseekmath} & 200 / 49.88 & 200 / 52.47 & 200 / 43.48 & 200 / 48.67 & 200 / 45.04 & 200 / 47.91 \\
SDPO~\citep{hubotter2026sdpo} & 150 / 8.62 & 150 / 0.18 & 145 / 11.17 & 125 / 39.84 & 110 / 31.07 & 136 / 18.18 \\
SRPO~\citep{li2026srpo} & 145 / 31.87 & 160 / 56.49 & 130 / 31.91 & 125 / 43.13 & 160 / 44.28 & 144 / 41.54 \\
\bottomrule
\end{tabular*}
\end{table}

\section{Case Study}
\label{sec:app-case-study}

To illustrate the weight allocation summarized in
Appendix~\ref{sec:token-weight-analysis},
Figure~\ref{fig:material-fec-token-weights} shows one matched pair of correct
and incorrect rollouts for a Materials question. Darker text groups mark greater mean acceptance under the local benefit and cost
proxies, not an estimated causal contribution to the answer. Uncolored
groups fall below the figure's coloring threshold, which does not establish that they are
irrelevant to reasoning. This example illustrates nonuniform allocation
within two responses rather than a general ordering of token importance.

\begin{figure}[H]
    \centering
    \includegraphics[
        width=\linewidth
    ]{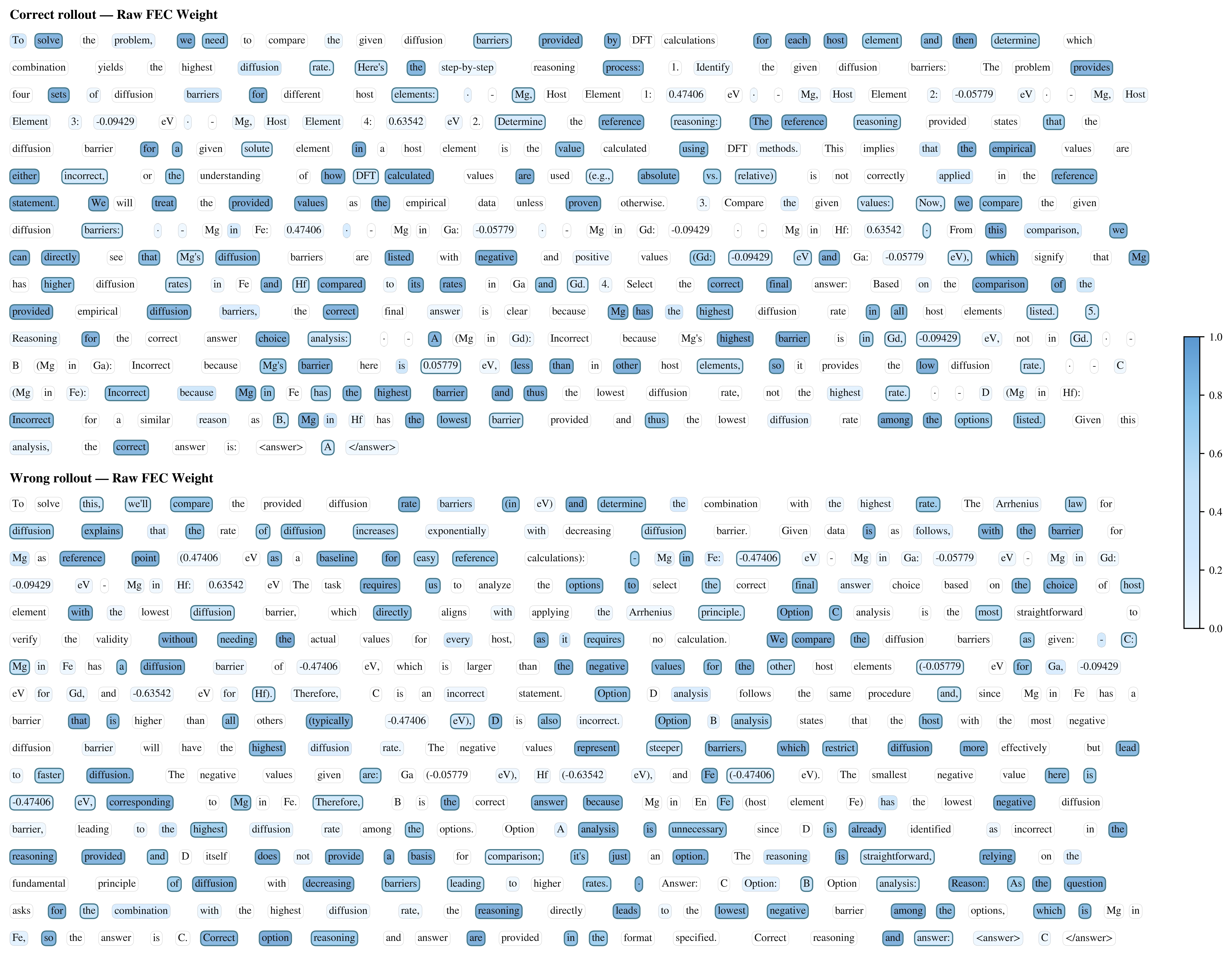}
    \caption{
        Token-level FEC weights for one matched pair of rollouts.
        The Materials question asks which host yields the highest diffusion rate
        for Mg, given diffusion barriers of $0.47406$ eV in Fe, $-0.05779$ eV
        in Ga, $-0.09429$ eV in Gd, and $0.63542$ eV in Hf. The ground-truth
        answer is A, corresponding to Mg in Gd. The upper correct rollout predicts A,
        whereas the lower incorrect rollout predicts C, corresponding to Mg in Fe.
        Adjacent subword pieces are grouped for readability.
        Fill color shows their mean offline scope-all raw FEC weight, with
        darker blue denoting a larger mean. A stronger outline marks a group
        containing a weight of at least 0.5. Groups whose maximum weight
        does not exceed $10^{-8}$ are left uncolored.
        Weights indicate controller acceptance, not causal contributions to
        correctness or the training-time routing frequency.
    }
    \label{fig:material-fec-token-weights}
\end{figure}

\end{document}